\documentclass[10pt,a4paper]{article}
\usepackage[numbers,square,comma,sort&compress]{natbib}%
\let\cite\citep
\usepackage{graphicx}%
\usepackage{multirow}%
\usepackage{amsmath,amssymb,amsfonts}%
\usepackage{amsthm}%
\usepackage{float}%
\usepackage{xcolor}%
\usepackage{textcomp}%
\usepackage{manyfoot}%
\usepackage{booktabs}%
\usepackage{algorithm}%
\usepackage{algorithmicx}%
\usepackage{algpseudocode}%
\usepackage{xspace}%
\usepackage{placeins}%
\usepackage{tikz}%
\usetikzlibrary{arrows.meta}%
\tikzset{every picture/.append style={color=black}}%

\usepackage[font=small,labelfont=bf]{caption}
\usepackage{microtype}
\usepackage[hidelinks,hypertexnames=false]{hyperref}

\theoremstyle{plain}%
\theoremstyle{remark}%
\theoremstyle{definition}%
\algnewcommand{\algorithmicoutput}{\textbf{Output:}}
\algnewcommand{\Output}{\item[\algorithmicoutput]}
\definecolor{algorithmgreen}{RGB}{0,102,68}
\newcommand{\OperatorComment}[1]{%
  \Statex \textcolor{algorithmgreen}{\textit{\texttt{//} #1}}}


\title{Chain of Operators: An Inference-Time Harness for In-Context Operator Learning}
\author{Minghui Yang\textsuperscript{1,2} \quad Chenghan Wu\textsuperscript{2} \quad
Ling Guo\textsuperscript{1} \quad Liu Yang\textsuperscript{2}\thanks{Corresponding author: \href{mailto:yangliu@nus.edu.sg}{yangliu@nus.edu.sg}}\\[5pt]
\small\textsuperscript{1}Department of Mathematics, Shanghai Normal University, China\\[3pt]
\small\textsuperscript{2}Department of Mathematics, National University of Singapore, Singapore}

\date{}
\begin{document}
\maketitle
\begin{abstract}

Pretrained in-context operator models can infer solution operators
from examples, yet their accuracy can deteriorate when a target problem lies
outside the distributions represented during training. We introduce Chain of
Operators (CHOP), an inference-time framework that composes calls to a frozen
In-Context Operator Network with explicit numerical operations. Task-specific
programs are selected on separate data and fixed before evaluation. Across
canonical partial differential equations and real-world forecasting, CHOP
improves direct model inference under several operator and distribution shifts,
although benefits vary across tasks and evaluation conditions. Selected modules
and programs can also remain effective without further evolution when reused with
another target mapping or frozen model. These results support explicit, modular
inference-time computation as a route to adapting in-context operator models while
keeping their parameters fixed.

\end{abstract}

\section{Introduction}

Many simulation and control problems can be expressed through solution operators,
which map inputs such as initial conditions, coefficient fields or costs to the
corresponding states. Classical numerical methods approximate this map by
discretizing and solving the governing equations for each new input, and repeated
solves can become prohibitively expensive at fine spatiotemporal resolutions.
Operator learning amortizes this cost by training a surrogate for the solution
operator from sampled input-output functions; after training, the surrogate
returns a solution for a new input through a single forward evaluation~\cite{kovachki2023neuraloperator,goswami2023physics,karniadakis2021physics}.
Architectures such as Deep Operator Networks (DeepONets)~\cite{lu2021deeponet}
and Fourier Neural Operators (FNOs)~\cite{li2020fourier} learn these
function-to-function mappings from data and can accelerate repeated simulations
by orders of magnitude when the target operator and inputs remain sufficiently
close to the training distribution~\cite{lu2022comprehensive}. Subsequent
developments have extended this paradigm with physics-informed
objectives~\cite{li2021physics,wang2021learning}, latent representations of
complex dynamics~\cite{kontolati2024learning}, general
geometries~\cite{li2022geofno} and alternative operator
architectures~\cite{raonic2023convolutional,tripura2023wavelet}. A frozen neural
operator, however, is generally most reliable over the operator families and
input distributions represented during training. Outside that support, errors
can increase, and recovering accuracy has commonly relied on parameter
retraining or fine-tuning~\cite{zhu2023reliable,goswami2022deep}.

Recent efforts have pre-trained neural operators across diverse physical systems
to amortize the cost of task-specific training~\cite{sun2024lemon,herde2024poseidon,hao2024dpot,rahman2024coda,subramanian2023towards,zhang2026deeponet}.
Such pre-training provides reusable representations and improves data efficiency.
Adapting these models to a genuinely new target operator, however, generally
still relies on downstream optimization or fine-tuning. A complementary
multi-operator strategy supplies the target physics through descriptors and
learns a conditional solution map~\cite{ye2024pdeformer,liu2024prosefd}. This
formulation can enable a single frozen model to serve multiple operators when
their descriptors lie within the learned support. It requires an accurate,
model-compatible description of the target physics, which may be unavailable
when a system is known only through observations.

In-Context Operator Networks (ICONs) provide an observation-based
alternative~\cite{liu2023incontext}. An ICON prompt pairs a small set of observed
input-output functions with a new query, allowing the frozen model to infer and
apply the demonstrated operator without its governing equation or gradient-based
adaptation. Subsequent work has broadened the training regimes, prompt content
and domains covered by this idea. Studies have explored language-augmented
prompts and unsupervised pre-training~\cite{yang2023prompting,chen2024dataefficient},
multiphysics and graph-valued spatiotemporal
data~\cite{cao2026vicon,wu2026graph}, and generative or probabilistic formulations
of parametric families and uncertainty~\cite{serrano2024zebra,kassai2026enma,zhang2025probabilistic}.
Applications now include optimal execution, transport on spaces of probability
measures and solid mechanics~\cite{meng2025solving,cole2026context,li2026context},
while theory has begun to characterize robustness and dependence on task
diversity~\cite{liu2023does,cole2024context}.
\begin{figure}[!htbp]
  \centering
  \includegraphics[width=\textwidth]{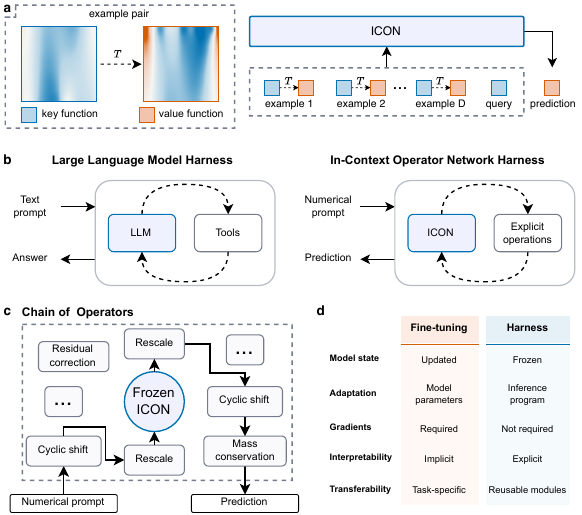}
  \caption{\textbf{Harnessing ICON with a Chain of Operators.}
    \textbf{a}, An in-context operator network (ICON) maps example input--output
    functions and a query to a prediction.
    \textbf{b}, Large language model (LLM) and ICON harnesses couple model calls
    to external tools or explicit numerical operations; dashed loops denote
    orchestration.
    \textbf{c}, Schematic Chain of Operators (CHOP) composition around a frozen
    ICON; the boxes denote modules used across the programs studied here.
    \textbf{d}, Conceptual comparison of fine-tuning and harness-based
    adaptation.}
  \label{fig:chop_main}
\end{figure}

Nevertheless, examples of a target operator do not guarantee that a frozen ICON
will infer and apply it reliably. Even a consistent set of examples may form a
numerical prompt outside the regime in which the model learned to identify and
apply operators. Prior work found that the same ICON can incur different errors
when a problem is written in mathematically equivalent
forms~\cite{yang2024pdegen}. This representation sensitivity motivates a more
general question: can an inference-time framework exploit ICON's prompt-level
flexibility to broaden the range of problems the frozen model can solve reliably,
without updating its weights?

A useful precedent comes from Large Language Models (LLMs). Parameter updates
remain a conventional route for adapting a pretrained LLM to a new
task~\cite{ding2023parameter,hu2023llm}. Frozen LLMs can also solve complex tasks
through external inference-time programs that coordinate intermediate reasoning
and tool calls~\cite{mialon2023augmented,wang2024survey,zhou2026externalization}.
Such programs are often called model harnesses: they decompose a task into
manageable steps while leaving the model parameters fixed. Comparable mechanisms
remain less developed for numerical foundation models, whose adaptation still
commonly uses gradient updates or learned auxiliary modules.

We therefore define an ICON harness as an external program that orchestrates
calls to a frozen ICON and explicit operations on numerical data
(Fig.~\ref{fig:chop_main}). By analogy with reasoning chains, we call our
construction Chain of Operators (CHOP). Its operators may transform prompts or
predictions, use held-out examples to estimate prompt-specific errors, and
incorporate structure supplied by the physical problem. Every operation and
dependency is explicit.
Because these choices define a large design space, we conduct a human-guided
search using the Evolving Ensemble of Agents (EvE) framework~\cite{Eve}, with
OpenAI Codex as its agent backend, to propose executable CHOP programs for each
evolution task.

We design and evaluate CHOP on two canonical PDE problems and a
real-world air-quality forecasting task~\cite{wu2026graph}. In each setting, the
frozen ICON is pretrained on a source problem family or region, while CHOP is
evolved on a designated task or transfer direction and fixed before evaluation on
held-out queries and targets. Across these settings, the evolved programs can
improve direct inference under changes in the target operator, the form and
spatial scale of the input--output functions, and the geographic domain, although
the gains vary across tasks and evaluation conditions. Reuse experiments further
show that selected components and complete programs can remain effective without
further evolution after changes to the target mapping, frozen model or equation
class. These results point toward an explicit and modular approach to adapting
in-context operator models through inference-time computation.

\section{Methods}

\subsection{Frozen ICON evaluation and CHOP programs}

An ICON prompt specifies a target operator through a set of input-output example pairs and queries its action on a new input. Let
$T:\mathcal{X}\to\mathcal{Y}$ denote the target operator and let
$\{(u_i,v_i)\}_{i=1}^{D}$ be example pairs satisfying $v_i=T(u_i)$. We refer to
$u_i$ and $v_i$ as the key and value of example $i$, respectively. For a new query $u_*$, let
$v_*=T(u_*)$ denote its target output. A single call to the frozen ICON on the original prompt approximates  $v_*$ as:
\begin{equation}
    \widehat v_{*,\mathrm{raw}}
    =\mathrm{ICON}\!\left(\{(u_i,v_i)\}_{i=1}^{D},u_*\right).
    \label{eq:icon}
\end{equation}
We call this direct evaluation on the unmodified prompt Raw ICON.

CHOP embeds calls to the frozen ICON within an explicit numerical program.
Starting from the original prompt, numerical operators may transform it, construct auxiliary or held-out prompts, and act on predictions
returned by the frozen ICON. Every ICON evaluation uses the same network.
Independent prompts may be evaluated together, while a prompt that depends on an
earlier prediction is evaluated only after that prediction becomes available. After
$Q$ logical ICON evaluations, CHOP applies the remaining numerical operators
to their predictions and returns $\widehat v_*$. These operators may invert a
prompt transformation, correct or project a prediction, select a route, or
combine candidates. The program therefore specifies both the numerical
operations and the dependencies among ICON evaluations.

CHOP draws on two sources of structure already available at inference time:
invertible transformations and observed example outputs. They motivate
induced-operator routes and in-context validation, respectively.

\subsection{Induced-operator routes}

The first general mechanism, an induced-operator route, changes the problem
presented to ICON while preserving a path back to the desired output. The example
pairs and $u_*$ may be used to choose invertible transformations $F$ and $G$.
Applying the same $F$ to every example input and the query, and the same $G$ to
every example output, gives
\begin{equation}
  G(v_i)
  =\left(G\circ T\circ F^{-1}\right)\!\left(F(u_i)\right).
  \label{eq:induced_operator_definition}
\end{equation}
\begingroup\sloppy
This identity determines the prediction problem seen by ICON. The transformed example pairs $
\{(F(u_i),G(v_i))\}_{i=1}^{D}$ describe the induced operator $
\Tilde{T}
=G\circ T\circ F^{-1}$. Its example inputs
are $F(u_i)$, its example outputs are $G(v_i)$, and its query is $F(u_*)$; the
corresponding target is therefore $\Tilde{T}(F(u_*))=G(T(u_*))$. ICON returns a prediction in these
transformed output coordinates, after which CHOP applies $G^{-1}$:
\par\endgroup
\begin{equation}
  \widehat v_*
  =G^{-1}\!\left(
    \mathrm{ICON}\!\left(
      \{(F(u_i),G(v_i))\}_{i=1}^{D},F(u_*)
    \right)
  \right).
  \label{eq:induced_operator_prediction}
\end{equation}
At the operator level, this route preserves the desired target: if ICON returns
$G(T(u_*))$, applying $G^{-1}$ recovers $T(u_*)$. The model approximation occurs
in the transformed ICON evaluation. Because a frozen ICON can incur different
errors on mathematically equivalent numerical prompts, $F$ and $G$ provide a way
to change that representation presented to ICON while retaining an exact map back to the physical
output.

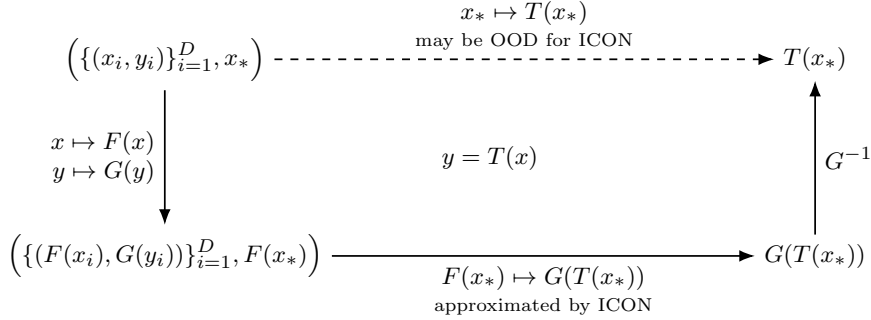
\begin{figure}[!htbp]
  \centering
  \begin{tikzpicture}[
      x=1cm,
      y=1cm,
      >=Latex,
      every node/.style={font=\small},
      route/.style={->, line width=0.65pt},
      target/.style={->, dashed, line width=0.65pt}
    ]
    \node (X)  at (0,2.6)
      {$\left(\{(u_i,v_i)\}_{i=1}^{D},u_*\right)$};
    \node (Xp) at (0,0)
      {$\left(\{(F(u_i),G(v_i))\}_{i=1}^{D},F(u_*)\right)$};
    \node (Yp) at (8.6,0) {$G(T(u_*))$};
    \node (Y)  at (8.6,2.6) {$T(u_*)$};
    \node at (4.3,1.3) {$v=T(u)$};

    \draw[target] (X) -- node[above, align=center]
      {$u_*\mapsto T(u_*)$\\[-0.2ex]
       {\footnotesize may be OOD for ICON}} (Y);
    \draw[route] (X) -- node[left, align=right]
      {$u\mapsto F(u)$\\$v\mapsto G(v)$} (Xp);
    \draw[route] (Xp) -- node[below, align=center]
      {$F(u_*)\mapsto G(T(u_*))$\\[-0.2ex]
       {\footnotesize approximated by ICON}} (Yp);
    \draw[route] (Yp) -- node[right] {$G^{-1}$} (Y);
  \end{tikzpicture}
  \caption{\textbf{Induced-operator route.}
    The dashed path is the desired map $u_*\mapsto T(u_*)$. The solid path
    transforms the prompt with $F$ and $G$, uses ICON to approximate the induced
    map $F(u_*)\mapsto G(T(u_*))$, and applies $G^{-1}$ to return the ICON
    prediction to the original output space.}
  \label{fig:induced_operator_route}
\end{figure}

Equations~\eqref{eq:induced_operator_definition} and
\eqref{eq:induced_operator_prediction} show the simplest form of this route.
Each prompt may define its own transformations. Within that prompt, the
transformed example pairs must describe one common operator, and the transformed
query must be posed to the same operator. CHOP(Conservation) generalizes this construction using time-indexed shifts. Translation equivariance ensures that example pairs shifted by different amounts still describe a common induced one-step operator, and that the shifted query is posed to that same operator.

VICON prompt normalization provides a concrete instance of this construction. It
computes channelwise means and standard deviations from the example inputs, then
uses the resulting affine normalization $\mathcal N$ for both input and output
functions~\cite{cao2026vicon}. Setting $F=G=\mathcal N$ presents ICON with
$\mathcal N\circ T\circ\mathcal N^{-1}$, and the inference procedure applies
$\mathcal N^{-1}$ to return the prediction to its physical scale.
\FloatBarrier
\subsection{In-context validation}

The second general mechanism is in-context validation. Its purpose is to use the
observed example outputs to reveal how a proposed inference procedure behaves on
the current prompt. Although the query target $v_*$ is unknown, every example contains a known output $v_h$. Predictions made by temporarily treating example inputs as queries can therefore be compared with their observed outputs.

\subsubsection{Residual correction}

Residual correction obtains such predictions from held-out prompts. For each
$h$, CHOP withholds pair $h$, treats $u_h$ as a temporary query and applies a
chosen prediction route using a prompt built from the remaining pairs. Let
$\widehat v_{h,\mathrm{base}}^{(-h)}$ denote the resulting prediction. The label
``base'' identifies the route whose query prediction may later be corrected, and
$(-h)$ records the withheld pair. Comparing this prediction with $v_h$ exposes the
residual
\begin{equation}
  R_h=v_h-\widehat v_{h,\mathrm{base}}^{(-h)},
  \qquad h=1,\ldots,D.
  \label{eq:residual_base_residuals}
\end{equation}
Applying the same route to the complete prompt gives
$\widehat v_{*,\mathrm{base}}$ at the actual query. The base route may be a direct
ICON evaluation, an induced-operator route or a longer fixed procedure.
The task-specific held-out constructions are given for CHOP(MFC) and CHOP(Air)
in Supplementary Algorithms~\ref{alg:chop_main} and~\ref{alg:air_chain}, respectively.

The held-out residuals are useful when errors observed at example inputs similar
to $u_*$ carry information about the error at the query. The module estimates
the query residual by giving larger
weight to residuals whose inputs are more similar to the query, then uses the
example pairs to calibrate the magnitude of that estimate.

ICON represents each function by values paired with coordinates, and the grids
may differ across examples and the query. The input $u_h$ must therefore be mapped
to the query grid before its similarity to $u_*$ is measured. The residual
$R_h$ must likewise be mapped to the prediction grid before it can contribute to
the correction.

To establish a common representation, we introduce two nearest-neighbor
interpolation operators. The map
$\mathcal I^u_{h\to *}$ interpolates the example input $u_h$ onto the query grid
for similarity evaluation, while
$\mathcal I^v_{h\to *}$ interpolates the
residual $R_h$ onto the prediction grid for residual correction. Each
operator reduces to the identity when its source and target grids
coincide. The implementation assigns each target point the value at the source
point with the smallest squared Euclidean coordinate distance; a tie selects the
first source point in the stored order.

The task-specific procedure also supplies a positive input scale $s_u$, computed
from the example pairs and query. The implementation divides input differences
by this scale before applying the numerical lower bound. After grid alignment,
the distance between example
input $u_h$ and query $u_*$ is
\begin{equation}
  d_{h*}
  =\operatorname{mean}\!\left[
    \left(
      \frac{\mathcal I^u_{h\to *}(u_h)-u_*}
           {s_u}
    \right)^2
  \right],
  \qquad
  \tau_*
  =\max\!\left\{\frac{1}{D}\sum_{h=1}^{D}d_{h*},\varepsilon_{\mathrm{res}}\right\},
  \label{eq:residual_transfer_distance}
\end{equation}
where $\tau_*$ sets the distance scale among the example inputs and
$\varepsilon_{\mathrm{res}}=10^{-6}$ is the numerical lower bound used by the
implementation. When the mean distance is larger than this bound, every
$d_{h*}/\tau_*$ and $d_{jh}/\tau_h$ is unchanged by the common scale $s_u$.
The scale affects the weights only when the numerical lower bound is active.
CHOP(MFC) supplies the full-prompt rescaling factor
$s_u=\sigma_{\mathrm{sc}}$ defined in Supplementary
Section~\ref{sec:supp_chop_mfc}.

Exponentially decaying weights give more influence to residuals whose inputs
are closer to the query. The resulting query correction is
\begin{equation}
  w_{h*}
  =\frac{\exp(-d_{h*}/\tau_*)}
         {\sum_{r=1}^{D}\exp(-d_{r*}/\tau_*)},
  \qquad
  \Delta_*
  =\sum_{h=1}^{D}w_{h*}\mathcal I^v_{h\to *}(R_h).
  \label{eq:residual_transfer}
\end{equation}
The estimate $\Delta_*$ gives the shape of a possible correction, but its full
magnitude may be too large. We calibrate that magnitude using the leave-one-out
residuals.
For each $h$, the residual $R_h$ is reconstructed from the other residuals,
\begin{equation}
  \Delta_h
  =\sum_{\substack{j=1\\j\ne h}}^{D}
    w_{jh}\mathcal I^v_{j\to h}(R_j),
  \qquad h=1,\ldots,D.
  \label{eq:residual_context_reconstruction}
\end{equation}
The weight $w_{jh}$ is computed from the similarity between $u_j$ and $u_h$, and
$\mathcal I^v_{j\to h}$ maps $R_j$ to the output grid of example $h$. The
construction of $\Delta_h$ omits $R_h$ and reuses every remaining residual exactly
as it was obtained from its own task-specific prompt. The pairs
$\{(\Delta_h,R_h)\}_{h=1}^{D}$ therefore calibrate the correction scale while
excluding the residual being reconstructed.

The implementation fits the correction coefficient in closed form and clips it
to the interval $[0,1]$:
\begin{equation}
\hat\alpha
=\operatorname{clip}_{[0,1]}\!\left(
  \frac{\sum_{h=1}^{D}\langle\Delta_h,R_h\rangle}
       {\max\!\left\{\sum_{h=1}^{D}\lVert\Delta_h\rVert_2^2,
                      \varepsilon_{\mathrm{res}}\right\}}
\right).
\label{eq:residual_scaling_coefficient}
\end{equation}
The inner product and norm range over all sampled output values.
The coefficient reflects how consistently the reconstructed residuals match the
observed leave-one-out residuals. The implementation applies the correction when
this fit reduces the mean squared residual error. Define
\begin{equation}
  E_{\mathrm{base}}
  =\frac{1}{D}\sum_{h=1}^{D}\operatorname{mean}(R_h^2),
  \qquad
  E_{\mathrm{fit}}
  =\frac{1}{D}\sum_{h=1}^{D}
    \operatorname{mean}\!\left[(R_h-\hat\alpha\Delta_h)^2\right],
  \label{eq:residual_gate_errors}
\end{equation}
and set $\alpha_{\mathrm{use}}=\hat\alpha$ when $E_{\mathrm{fit}}$ is finite and
$E_{\mathrm{fit}}<E_{\mathrm{base}}$. Every other case sets
$\alpha_{\mathrm{use}}=0$.
The corrected branch prediction is
\begin{equation}
\widehat v_{*,\mathrm{corr}}
=\widehat v_{*,\mathrm{base}}
+\alpha_{\mathrm{use}}\Delta_*.
\label{eq:residual_corrected_prediction}
\end{equation}

Algorithm~\ref{alg:chop_G} receives the base predictions and input scale, then
returns both the corrected prediction and $E_{\mathrm{fit}}$. The returned
error is computed with $\hat\alpha$ even when the internal gate sets
$\alpha_{\mathrm{use}}=0$, because CHOP(MFC) later uses this pre-gate error to
compare its complete rescaled and Raw ICON routes.
\begin{algorithm}[!ht]
\caption{Residual-correction operator $G_{\mathrm{corr}}$.}
\label{alg:chop_G}
\begin{algorithmic}[1]
\small
\Require $D\geq2$ pairs $\{(u_i,v_i)\}_{i=1}^{D}$, query $u_*$,
  base predictions $\{\widehat v_{h,\mathrm{base}}^{(-h)}\}_{h=1}^{D}$,
  each sampled on the output grid of $v_h$ and obtained from a task-specific
  prompt that excludes pair $h$, full-prompt prediction
  $\widehat v_{*,\mathrm{base}}$ on the query-output grid and positive input
  scale $s_u$.
\Output Corrected prediction $\widehat v_{*,\mathrm{corr}}$ and fitted
  residual error $E_{\mathrm{fit}}$.

\OperatorComment{Compute the held-out base residuals}
\For{$h=1,\ldots,D$}
  \State $R_h\leftarrow v_h-\widehat v_{h,\mathrm{base}}^{(-h)}$.
\EndFor

\OperatorComment{Transfer the residuals to the query}
\For{$h=1,\ldots,D$}
  \State $d_{h*}\leftarrow
    \operatorname{mean}[((\mathcal I^u_{h\to *}(u_h)-u_*)/
    s_u)^2]$.
\EndFor
\State $\tau_*\leftarrow
  \max\{D^{-1}\sum_{h=1}^{D}d_{h*},\varepsilon_{\mathrm{res}}\}$.
\State $w_{h*}\leftarrow
  \exp(-d_{h*}/\tau_*)/
  \sum_{r=1}^{D}\exp(-d_{r*}/\tau_*)$ for each $h$.
\State $\Delta_*\leftarrow
  \sum_{h=1}^{D}w_{h*}\mathcal I^v_{h\to *}(R_h)$.

\OperatorComment{Reconstruct each held-out residual from the remaining residuals}
\For{$h=1,\ldots,D$}
  \For{$j=1,\ldots,D$ with $j\ne h$}
    \State $d_{jh}\leftarrow
      \operatorname{mean}[((\mathcal I^u_{j\to h}(u_j)-u_h)/
      s_u)^2]$.
  \EndFor
  \State $\tau_h\leftarrow
    \max\{(D-1)^{-1}\sum_{j\ne h}d_{jh},\varepsilon_{\mathrm{res}}\}$.
  \State $w_{jh}\leftarrow
    \exp(-d_{jh}/\tau_h)/
    \sum_{r\ne h}\exp(-d_{rh}/\tau_h)$ for each $j\ne h$.
  \State $\Delta_h\leftarrow
    \sum_{j\ne h}w_{jh}\mathcal I^v_{j\to h}(R_j)$.
\EndFor

\OperatorComment{Fit the correction magnitude and apply the internal gate}
\State Compute $\hat\alpha$ from
  equation~\eqref{eq:residual_scaling_coefficient}.
\State Compute $E_{\mathrm{base}}$ and $E_{\mathrm{fit}}$ from
  equation~\eqref{eq:residual_gate_errors}.
\If{$E_{\mathrm{fit}}$ is finite and $E_{\mathrm{fit}}<E_{\mathrm{base}}$}
  \State $\alpha_{\mathrm{use}}\leftarrow\hat\alpha$.
\Else
  \State $\alpha_{\mathrm{use}}\leftarrow0$.
\EndIf
\State $\widehat v_{*,\mathrm{corr}}\leftarrow
  \widehat v_{*,\mathrm{base}}+\alpha_{\mathrm{use}}\Delta_*$.
\State \Return $\widehat v_{*,\mathrm{corr}},E_{\mathrm{fit}}$.
\end{algorithmic}
\end{algorithm}

\subsubsection{Combining candidate predictions}

A task-specific program may produce several plausible predictions whose
relative accuracy changes from one prompt to another. Their query errors are
unavailable because $v_*$ is unknown. The observed example outputs reveal
both the error of each candidate and whether two candidates tend to err in the
same direction. Candidate combination uses this joint error information to fit
a convex mixture. A delete-one stability estimate determines how far its
weights move from a designated reference prediction.

For $D\geq2$ examples, suppose the task-specific program supplies $K$ predictions
$\{\widehat v_{h,k}\}_{k=1}^{K}$ at every example input and corresponding
predictions $\{\widehat v_{*,k}\}_{k=1}^{K}$ at the query. For each $h$, the
candidates at $u_h$ are sampled on the output grid of $v_h$; all query
candidates share the query-output grid. Candidate 1 is the reference. Since
$v_h$ is observed, define the candidate errors and their matrix contributions
by
\begin{equation}
  e_{h,k}=\widehat v_{h,k}-v_h,
  \qquad
  (M_h)_{kr}=\langle e_{h,k},e_{h,r}\rangle_h,
  \qquad
  M=\frac{1}{D}\sum_{h=1}^{D}M_h,
  \label{eq:candidate_error_matrix}
\end{equation}
where $\langle u,v\rangle_h=N_h^{-1}\sum_{q=1}^{N_h}u_qv_q$ is the average of
the pointwise products over the $N_h$ grid points of example $h$. The diagonal
entries of $M$ measure the individual mean squared errors, whereas the
off-diagonal entries are cross-moments between the errors of different
candidates. Define the probability simplex
\begin{equation}
  \Delta_K
  =\left\{a\in\mathbb R^K:a_k\geq0,\ \sum_{k=1}^{K}a_k=1\right\}.
  \label{eq:candidate_weight_simplex}
\end{equation}
For $a\in\Delta_K$, the mean squared error of the combined prediction across
the example inputs is $a^{\top}Ma$.
We therefore estimate the combination weights by
\begin{equation}
  \hat a
  =\underset{a\in\Delta_K}{\arg\min}\;a^{\top}Ma.
  \label{eq:candidate_weight_fit}
\end{equation}
Because $M$ is positive semidefinite, the resulting optimization problem is
convex. With only $D$ matrix contributions, however, $\hat a$ may depend
strongly on one example. We measure this sensitivity with a delete-one
jackknife. For each $h$, we keep the supplied candidate rows fixed, omit $M_h$
and solve the same convex problem using the remaining contributions:

\begin{equation}
  \begin{aligned}
    M^{(-h)}
    &=\frac{1}{D-1}\sum_{j\ne h}M_j,\\
    \hat a^{(-h)}
    &=\underset{a\in\Delta_K}{\arg\min}\;
      a^{\top}M^{(-h)}a.
  \end{aligned}
  \label{eq:candidate_weight_jackknife_fit}
\end{equation}
Let $\bar a=D^{-1}\sum_{h=1}^{D}\hat a^{(-h)}$. The resulting delete-one
jackknife covariance is
\begin{equation}
  V
  =\frac{D-1}{D}\sum_{h=1}^{D}
    \left(\hat a^{(-h)}-\bar a\right)
    \left(\hat a^{(-h)}-\bar a\right)^{\top}.
  \label{eq:candidate_weight_jackknife_covariance}
\end{equation}
The matrix $V$ records how much the fitted weights change across these
delete-one jackknife refits. Let
$a_{\mathrm{ref}}=(1,0,\ldots,0)^{\top}$ select the designated reference
candidate and $\delta=\hat a-a_{\mathrm{ref}}$. We stabilize the fitted weights
by
\begin{equation}
  \lambda
  =\frac{\delta^{\top}M\delta}
         {\delta^{\top}M\delta+\operatorname{tr}(MV)},
  \qquad
  a=a_{\mathrm{ref}}+\lambda\delta.
  \label{eq:candidate_weight_shrinkage}
\end{equation}
We set $\lambda=0$ unless the denominator is positive.
The term $\delta^{\top}M\delta$ measures the displacement from the reference
weights through the quadratic form defined by $M$. The term
$\operatorname{tr}(MV)$ measures the variation among the delete-one jackknife
refits in the same quadratic form. Through their ratio, $\lambda$ retains more
of the fitted displacement when the delete-one refits are stable and shrinks
the weights toward $a_{\mathrm{ref}}$ as their variation increases. Because
$0\leq\lambda\leq1$, the stabilized weights remain
in $\Delta_K$.

After stabilizing the weights using the examples, we apply them to the $K$
candidate predictions at the query,
\begin{equation}
  \widehat v_*
  =\sum_{k=1}^{K}a_k\widehat v_{*,k}.
  \label{eq:candidate_combination}
\end{equation}
Changing the candidate set
changes $K$ and $M$, while the fitting and composition remain the same.
CHOP(Air) uses Raw ICON as its reference candidate and estimates a separate
weight vector for PM$_{2.5}$ and O$_3$.

Algorithm~\ref{alg:context_validated_combination} carries the candidate errors
through the simplex fit, delete-one jackknife stabilization and final query
combination.

\begin{algorithm}[!ht]
\caption{Candidate combination.}
\label{alg:context_validated_combination}
\begin{algorithmic}[1]
\small
\Require $D\geq2$ observed outputs $\{v_h\}_{h=1}^{D}$, $K\geq1$ supplied
  predictions $\{\widehat v_{h,k}:h=1,\ldots,D,\ k=1,\ldots,K\}$ and
  $\{\widehat v_{*,k}\}_{k=1}^{K}$, with the candidates at $u_h$ sampled on
  the output grid of $v_h$, all query candidates sampled on one output grid,
  and designated reference candidate $k=1$.
\Output Combined query prediction $\widehat v_*$.

\OperatorComment{Candidate errors: form one matrix contribution per example}
\For{$h=1,\ldots,D$}
  \State $e_{h,k}\leftarrow\widehat v_{h,k}-v_h$ for all $k$.
  \State Form $M_h\in\mathbb R^{K\times K}$ with
    $(M_h)_{kr}\leftarrow\langle e_{h,k},e_{h,r}\rangle_h$ for all $k,r$.
\EndFor
\State $M\leftarrow D^{-1}\sum_{h=1}^{D}M_h$.

\OperatorComment{Simplex fit: estimate the candidate weights}
\State $\hat a\leftarrow
  \underset{a\in\Delta_K}{\arg\min}\;a^{\top}Ma$.

\OperatorComment{Delete-one jackknife: stabilize the fitted weights}
\For{$h=1,\ldots,D$}
  \State $M^{(-h)}\leftarrow(D-1)^{-1}\sum_{j\ne h}M_j$.
  \State $\hat a^{(-h)}\leftarrow
    \underset{a\in\Delta_K}{\arg\min}\;
    a^{\top}M^{(-h)}a$.
\EndFor
\State $\bar a\leftarrow D^{-1}\sum_{h=1}^{D}\hat a^{(-h)}$.
\State $V\leftarrow\dfrac{D-1}{D}\sum_{h=1}^{D}
  (\hat a^{(-h)}-\bar a)(\hat a^{(-h)}-\bar a)^{\top}$.
\State $a_{\mathrm{ref}}\leftarrow(1,0,\ldots,0)^{\top}$;
  $\delta\leftarrow\hat a-a_{\mathrm{ref}}$.
\If{$\delta^{\top}M\delta+\operatorname{tr}(MV)>0$}
  \State $\lambda\leftarrow
    \dfrac{\delta^{\top}M\delta}
    {\delta^{\top}M\delta+\operatorname{tr}(MV)}$.
\Else
  \State $\lambda\leftarrow0$.
\EndIf
\State $a\leftarrow a_{\mathrm{ref}}+\lambda\delta$.

\OperatorComment{Combination: apply the stabilized weights to the query candidates}
\State \Return $\widehat v_*\leftarrow
  \sum_{k=1}^{K}a_k\widehat v_{*,k}$.
\end{algorithmic}
\end{algorithm}

Residual correction estimates an additive error for one base route, whereas
candidate combination fits a convex mixture of several predictions. Both use
observed example outputs to evaluate inference choices that can then be applied at the
unknown query. The induced-operator route serves a different role by changing
the numerical representation presented to ICON.

\subsection{Physics-derived operators}

CHOP can also incorporate numerical operators justified by the task's physics. A symmetry can
justify a coordinate transformation around the ICON call, while a known invariant
can define an output projection. In the conservation-law problem, the moving-frame
shift follows from translation symmetry, while a mean-adjustment operator enforces
mass conservation by matching the predicted mean to that of the query.

\subsection{Program evolution and evaluation}

Together, induced-operator routes, in-context validation and physically
justified operators define a structured space of executable CHOP programs. A candidate program specifies the sequence of numerical operations, the ICON evaluations to be performed, the dependencies among them, and the rule that produces the final prediction. We search this program space with Evolving Ensemble of
Agents (EvE)~\cite{Eve}.

We run EvE with OpenAI Codex as its agent backend, accessed through a
non-interactive Codex CLI. Researchers define the task, operator library, physical
constraints and evaluation objective; these choices delimit the programs that
Codex can propose. During each iteration, EvE orchestrates Codex calls that revise
a population of executable programs, and a task evaluator scores them on an
evolution set reserved for program development. Prediction error determines which
candidates remain, while the ICON weights stay fixed. Evolution therefore changes
the inference-time program while preserving the underlying model.

The first candidates produced by EvE with Codex were executable, but their
complexity obscured the roles and dependencies of individual operations.
After the EvE search, researchers and Codex agents worked together to revise and
simplify these candidates until they could be inspected directly.
Every performance signal that guided this process came from the corresponding
evolution set. All revisions and selection decisions were completed before the
resulting program was run on any designated evaluation dataset. Because agent
proposals and human judgement both shaped this process, repeating it need not
produce identical CHOP source code. After the final program for a task was fixed,
it was run once on each designated evaluation dataset. Neither the evaluation
targets nor the resulting metrics were returned to the search or used for later
revision or tuning. The fixed program sources, evaluation pipeline, frozen model
checkpoints, evaluation data and recorded settings allow the reported evaluation
outputs and metrics to be regenerated.

Evaluation queries and targets are disjoint from the evolution set and include operators, parameter regimes,
horizons or regions that evolution did not encounter. For each test prompt, the fixed
program derives every prompt-dependent quantity from the example pairs and
$u_*$; the unknown target $v_*$ is used only to score the completed prediction.
The Supplementary Methods give the complete definitions of the selected
programs, including task-specific branches and numerical safeguards.

\FloatBarrier

\section{Results}
Raw ICON evaluates the original prompt with the frozen model. CHOP instead
arranges calls to the same model in a Chain of Operators: explicit operations
transform prompts or predictions, assess routes against known example outputs,
impose physical structure and combine predictions. The chain fixes the
dependencies among calls. Methods introduce the general mechanisms;
Supplementary Methods specify each program.

The programs reported below were developed through a human-guided search.
Researchers specified the broad framework of induced-operator routes and
in-context validation against observed example outputs, together with
task-specific physical constraints and evaluation criteria. Within this space,
EvE, with OpenAI Codex as its agent backend, produced initial executable
programs. Because their complexity obscured the roles and dependencies of
individual operations, researchers and Codex agents jointly simplified them
after the search. All performance evidence used for search, simplification and
selection came from the corresponding evolution set. Each final program was
fixed before being run once on each designated evaluation dataset; neither
evaluation targets nor the resulting metrics informed subsequent revision or
tuning.

We first study two PDE settings with complementary distribution shifts.
Conservation laws test repeated propagation under fluxes outside the pretraining
family, where one-step errors can accumulate. Mean-field control (MFC) tests
single-shot mappings as shorter correlation lengths introduce spatial variation
below the pretraining scale. Together, they test long autoregressive rollouts and
the reconstruction of sharp solutions.

\subsection{Conservation laws beyond the pretraining flux family}

Conservation laws describe transport and shock formation, with the flux function
determining how the solution propagates. We consider the one-dimensional scalar conservation law
\refstepcounter{problem}
\begin{equation}
  \makebox[0pt][r]{\textbf{Problem \theproblem}\qquad}
  \partial_t u(x,t)+\partial_x f(u(x,t))=0,
  \qquad x\in[0,1],
  \label{eq:consv_pde}
\end{equation}
with periodic boundary conditions. Its time-$\Delta t$ solution operator,
$S_{f,\Delta t}:u(\cdot,t)\mapsto u(\cdot,t+\Delta t)$, advances the state by
$\Delta t$ and changes with the choice of flux $f$. The initial numerical prompt
contains $D=5$ consecutive one-step transitions from a single trajectory,
$(u_i,v_i)=(u(\cdot,t_i),u(\cdot,t_i+\Delta t))$, where
$t_i=t_1+(i-1)\Delta t$. The final example output is also the first query,
$u_*=u(\cdot,t_1+D\Delta t)=v_D$, whose target is
$v_*=u(\cdot,t_1+(D+1)\Delta t)$. During autoregressive rollout, the five
example transitions remain fixed, while each prediction becomes the query for
the next step.

The frozen ICON is pretrained only on cubic fluxes $f(u)=au^3+bu^2+cu$, with
$(a,b,c)\in[-1,1]^3$. We evaluate it on
three qualitatively different fluxes outside this family: the smooth periodic flux
$f(u)=\sin u-\cos u$, the saturating flux $f(u)=\tanh u$, and the
Buckley--Leverett flux $f(u)=u^2/\bigl(u^2+(1-u)^2\bigr)$ for two-phase flow in
porous media. Evolution uses 100 trajectories from the first flux. Evaluation
uses 500 independent trajectories for each of the three fluxes. The other two
fluxes are absent from pretraining and evolution, providing a transfer test.

\begin{figure}[!t]
  \centering
  \includegraphics[width=\textwidth]{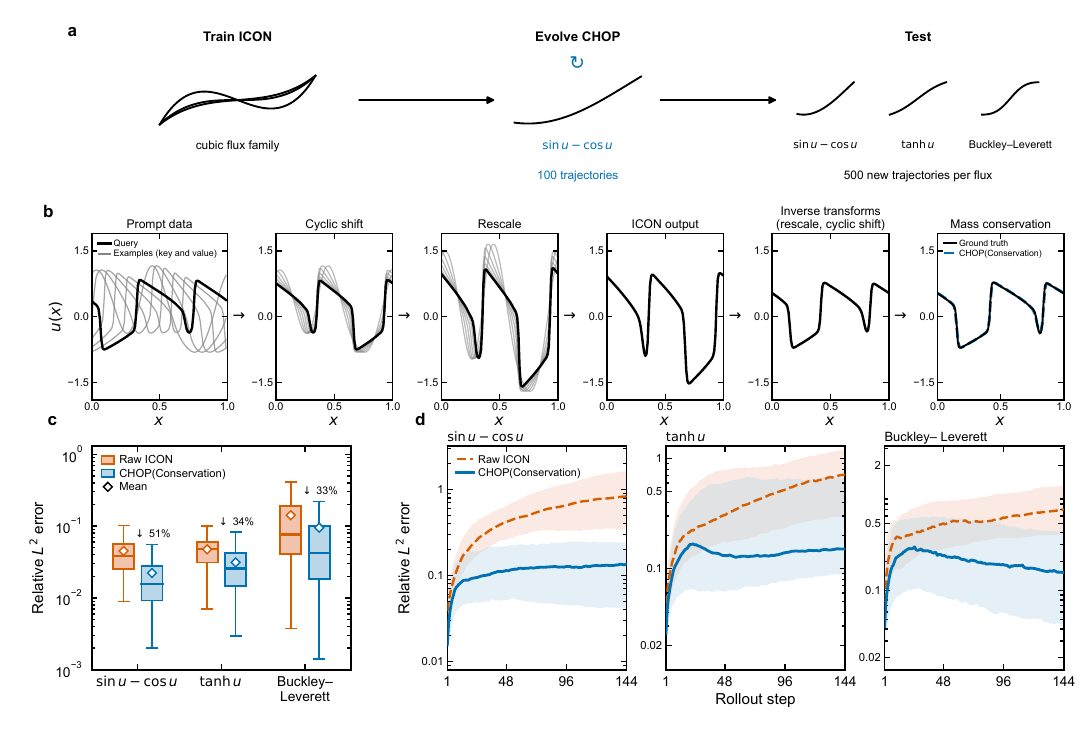}
  \caption{\textbf{CHOP improves prediction accuracy throughout autoregressive rollouts for
  non-cubic conservation-law fluxes.}
  \textbf{a}, ICON pretraining on cubic fluxes, CHOP evolution on
  $\sin u-\cos u$ and testing on three non-cubic fluxes.
  \textbf{b}, An illustrative program trace in which CHOP(Conservation)
  transforms the prompt, evaluates ICON, then applies the inverse transformation
  and enforces mass conservation.
  \textbf{c}, Raw ICON and CHOP(Conservation) single-step relative $L^2$ errors;
  percentages give reductions in the arithmetic mean from Raw ICON.
  \textbf{d}, Median relative $L^2$ error across the 144-step autoregressive
  rollout, with shading spanning the interquartile range at each step. Each flux
  has $n=500$ independent test trajectories in \textbf{c} and \textbf{d}. In
  \textbf{c}, boxes span the interquartile range, centre lines mark medians,
  diamonds mark arithmetic means and whiskers extend to the most extreme values
  within $1.5\times$ the interquartile range; observations beyond the whiskers
  are not shown.}
  \label{fig:conservation_results}
\end{figure}

A conservation-law prompt exposes three structures that can be handled
explicitly: periodic transport, prompt-dependent numerical ranges and conservation
of the query mean. CHOP(Conservation) assigns one operation to each role in a
single-evaluation program ($Q=1$). The shift stage estimates an integer displacement
from the most recent example pair and uses its time-indexed multiples to place the
visible fields in a moving frame; one pooled affine transformation then rescales the
shifted prompt. The complete program is
\begin{equation}
  \begin{aligned}
    \left(\{(u_i,v_i)\}_{i=1}^{D},u_*\right)
    &\xrightarrow{\mathrm{shift}}
    \left(\{(u_i^s,v_i^s)\}_{i=1}^{D},u_*^s\right)
    \\
    &\xrightarrow{\mathrm{rescale}}
    \left(\{(u_i^{\mathrm{sc}},v_i^{\mathrm{sc}})\}_{i=1}^{D},
      u_*^{\mathrm{sc}}\right)
    \xrightarrow{\mathrm{ICON}} \widehat v^{\mathrm{sc}},\\
    \widehat v^{\mathrm{sc}}
    &\xrightarrow{\mathrm{inverse\ rescale}} \widehat v_s
    \xrightarrow{\mathrm{inverse\ shift}} \widehat v
    \xrightarrow{\mathrm{mass\ conservation}} \widehat v_*.
  \end{aligned}
  \label{eq:chop_conservation}
\end{equation}
The shift and rescaling stages realize the input- and output-side transformations
in Fig.~\ref{fig:induced_operator_route} at the appropriate temporal indices. By
translation equivariance, the transformed example pairs describe one common
induced one-step operator, to which the transformed query is posed.
Supplementary Algorithm~\ref{alg:consv_chain_eval} gives the exact transforms
and numerical safeguards. CHOP introduces only prompt- or law-derived quantities;
it adds no flux-function input or learned parameters
(Fig.~\ref{fig:conservation_results}b).

Across the three fluxes, CHOP reduces arithmetic mean
single-step relative $L^2$ error by 33--51\%
(Fig.~\ref{fig:conservation_results}c), with improvements extending to the two
fluxes excluded from evolution. The selected program therefore transfers beyond
the flux used for evolution, and its lower error persists through repeated
autoregressive propagation: the median error remains lower at every one of the
144 steps (Fig.~\ref{fig:conservation_results}d), and the final-step arithmetic
mean error is reduced by 47--90\% (Table~\ref{tab:consv_singlestep}).

\begin{table}[!htbp]
  \centering
  \caption{\textbf{Conservation-law relative $L^2$ error.}
    Entries are mean $\pm$ sample s.d.\ over $n=500$ paired trajectories for
    each flux and evaluation point.
    Reductions are relative to Raw ICON.}
  \label{tab:consv_singlestep}
  \small
  \setlength{\tabcolsep}{2.5pt}
  \begin{tabular}{llccc}
    \toprule
    Flux $f(u)$ & Evaluation & Raw ICON & CHOP(Conservation) & Reduction \\
    \midrule
    \multirow{2}{*}{$\sin u-\cos u$}
      & Single step & $0.0451\pm0.0312$ & $0.0222\pm0.0219$ & 50.68\% \\
      & Step 144 & $2.2413\pm4.7878$ & $0.2186\pm0.3567$ & 90.25\% \\
    \midrule
    \multirow{2}{*}{$\tanh u$}
      & Single step & $0.0475\pm0.0199$ & $0.0313\pm0.0223$ & 34.16\% \\
      & Step 144 & $0.8105\pm0.6017$ & $0.4325\pm0.5390$ & 46.64\% \\
    \midrule
    \multirow{2}{*}{$u^2/(u^2+(1-u)^2)$}
      & Single step & $0.1418\pm0.1530$ & $0.0951\pm0.1307$ & 32.93\% \\
      & Step 144 & $2.1342\pm5.7115$ & $0.2539\pm0.2939$ & 88.10\% \\
    \bottomrule
  \end{tabular}
\end{table}

\subsection{Mean-field control across operator families and spatial scales}

Our second setting changes both the prediction mode and the
source of distribution shift. Mean-field control describes the optimal steering
of a population density, with applications including crowd motion and resource
allocation. We consider the transport and terminal-cost objective
\refstepcounter{problem}
\begin{equation}
\label{eq:mfc}
  \makebox[0pt][r]{\textbf{Problem \theproblem}\qquad}
  \inf_{\rho,m}
  \int_0^1\!\!\int_0^1\frac{c\,m^2(t,x)}{2\,\rho(t,x)}\,dx\,dt
  +\int_0^1 g(x)\,\rho(1,x)\,dx,
\end{equation}
subject to the Fokker--Planck constraint
\begin{equation}
  \partial_t\rho+\partial_x m=\mu\,\partial_{xx}\rho,
\end{equation}
with periodic boundary conditions in $x$ and a prescribed initial density
$\rho(0,x)$.

The five MFC operator families separate two kinds of transfer.
Three families vary the input and output dimensions while mapping
densities to densities; two instead take a terminal cost as input and return a
density. Each prompt holds the function named in the subscript fixed, so its
examples and query belong to one operator instance.

\begin{figure}[!htbp]
  \centering
  \includegraphics[width=0.86\textwidth]{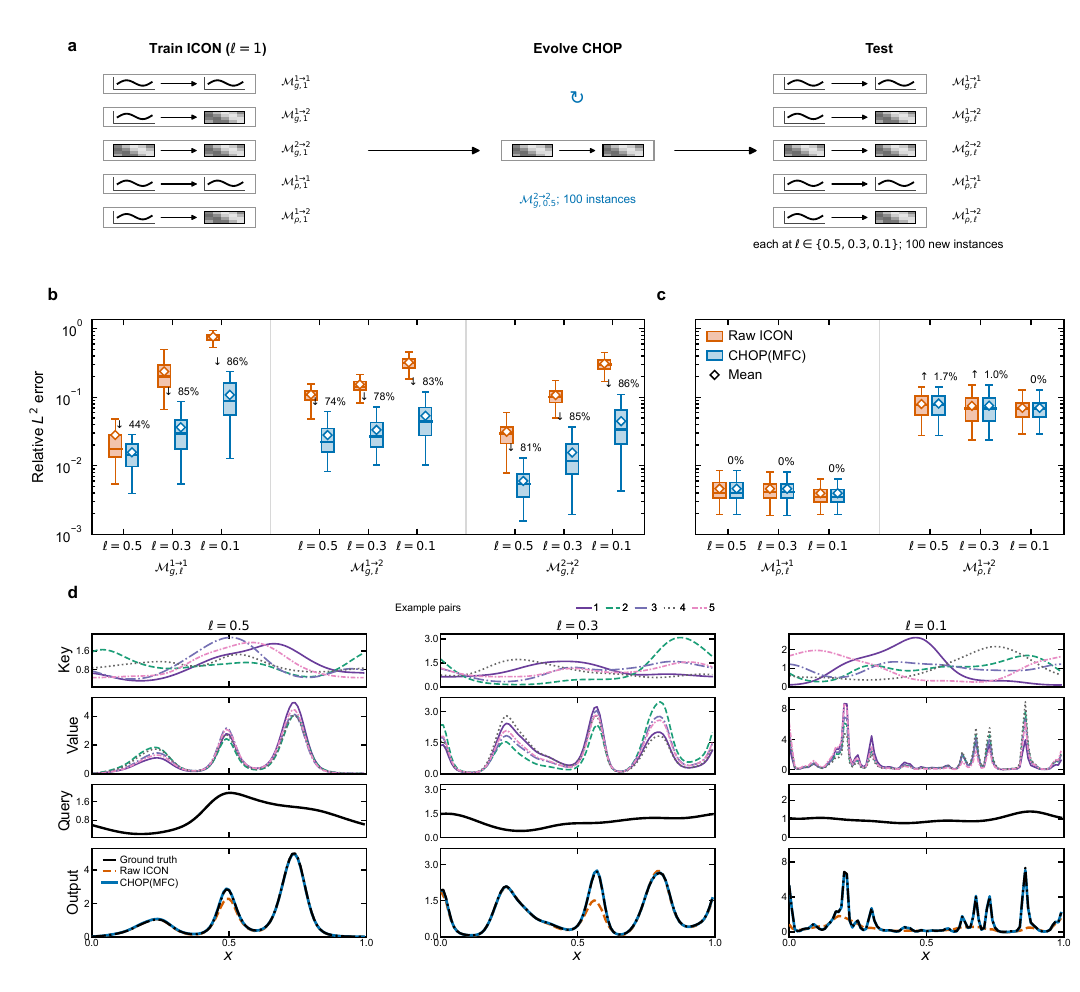}
  \caption{\textbf{CHOP(MFC) performs differently on
  density-to-density and cost-to-density
  operator families.}
  \textbf{a}, Mean-field-control pretraining, CHOP evolution and test routes.
  Line and field glyphs denote one- and two-dimensional functions, respectively,
  and each outline encloses one input--output mapping.
  \textbf{b}, Relative $L^2$ error for the three
  density-to-density
  operator families.
  \textbf{c}, Relative $L^2$ error for the two
  cost-to-density
  families.
  Percentages in \textbf{b} and \textbf{c} report arithmetic-mean changes from
  Raw ICON; downward and upward arrows denote decreases and increases. Each box
  contains $n=100$ independently sampled operator instances. Boxes span the
  interquartile range, centre lines mark medians, diamonds mark arithmetic means
  and whiskers extend to the most extreme values within $1.5\times$ the
  interquartile range; observations beyond the whiskers are not shown.
  \textbf{d}, Density-to-density prompts and predictions at the three displayed
  length scales; the examples and query in each column share one operator.}
  \label{fig:mfc_results}
\end{figure}

When a sampled terminal cost $g$ is fixed, the examples vary either the initial
density or the early part of the density trajectory. This gives
$\mathcal{M}_{g,\ell}^{1\to1}:\rho(0,x)\mapsto\rho(1,x)$,
$\mathcal{M}_{g,\ell}^{1\to2}:\rho(0,x)\mapsto\rho_{[1/2,1]}(t,x)$ and
$\mathcal{M}_{g,\ell}^{2\to2}:\rho_{[0,1/2]}(t,x)\mapsto
\rho_{[1/2,1]}(t,x)$. When the sampled initial density is fixed instead, the
examples vary $g$, producing
$\mathcal{M}_{\rho,\ell}^{1\to1}:g(x)\mapsto\rho(1,x)$ and
$\mathcal{M}_{\rho,\ell}^{1\to2}:g(x)\mapsto\rho_{[1/2,1]}(t,x)$.
The superscript records the dimensions of the input and output domains, and
$\rho_{[a,b]}$ denotes the density trajectory restricted to $t\in[a,b]$.
The correlation length $\ell$ controls the spatial variation of the sampled
terminal
cost or initial density; decreasing $\ell$ shifts the operator distribution
toward finer spatial structure.

Each prompt draws six distinct key--value pairs without replacement from one
operator instance, using five as examples and only the sixth key as the query.
ICON is pretrained on all five families at $\ell=1$.
CHOP(MFC) is evolved on 100 independently sampled instances of
$\mathcal{M}_{g,0.5}^{2\to2}$ and fixed before evaluation on 100 new instances
for each family and $\ell\in\{0.5,0.3,0.1\}$. Fourteen of the fifteen conditions
are absent from evolution, testing transfer across input--output geometry, input
role and spatial scale.

CHOP(MFC) uses leave-one-out errors to make two decisions around a prompt-derived
affine rescaling. Within each complete or held-out prompt, all visible example
inputs, example outputs and the query input determine one prompt-specific affine
centre and scale. Withholding each example output supplies errors for the
uncorrected rescaled route ($E_{\mathrm{base}}$), the pre-gate fitted-correction
error $E_{\mathrm{fit}}$ for $G_{\mathrm{corr}}$, and Raw ICON
($E_{\mathrm{raw}}$). The correction is retained when $E_{\mathrm{fit}}$ is
finite and below
$E_{\mathrm{base}}$; the rescaled route is selected when the same pre-gate
$E_{\mathrm{fit}}$ is finite and below $E_{\mathrm{raw}}$. The complete routes are
\begin{equation}
  \begin{aligned}
    \left(\{(u_i,v_i)\}_{i=1}^{D},u_*\right)
    &\xrightarrow{\mathrm{rescale}}
    \left(\{(u_i^{\mathrm{sc}},v_i^{\mathrm{sc}})\}_{i=1}^{D},
      u_*^{\mathrm{sc}}\right)
    \xrightarrow{\mathrm{ICON}} \widehat v^{\mathrm{sc}},\\
    \widehat v^{\mathrm{sc}}
    &\xrightarrow{\mathrm{inverse\ rescale}} \widehat v_{*,\mathrm{base}}
    \xrightarrow{G_{\mathrm{corr}}}
      \widehat v_{*,\mathrm{corr}},\\
    \left(\{(u_i,v_i)\}_{i=1}^{D},u_*\right)
    &\xrightarrow{\mathrm{Raw\ ICON}} \widehat v_{*,\mathrm{raw}},\\
    \widehat v_{*,\mathrm{MFC}}
    &=\begin{cases}
      \widehat v_{*,\mathrm{corr}},
        & E_{\mathrm{fit}}\text{ finite and }E_{\mathrm{fit}}<E_{\mathrm{raw}},\\
      \widehat v_{*,\mathrm{raw}}, & \text{otherwise}.
    \end{cases}
  \end{aligned}
\end{equation}
If the internal gate rejects the correction, the first branch is the uncorrected
rescaled prediction. Algorithm~\ref{alg:chop_G} and Appendix
Algorithm~\ref{alg:chop_main} specify residual transfer and numerical safeguards.

\begin{table}[htbp]
  \centering
  \caption{\textbf{Mean-field-control relative $L^2$ error.}
    Entries are mean $\pm$ sample s.d.\ over $n=100$ independently sampled
    operator instances per task. Reductions are relative to Raw ICON; negative
    values indicate increases.}
  \label{tab:main}
  \small
  \setlength{\tabcolsep}{2.5pt}
  \begin{tabular}{l c ccc}
    \toprule
    Operator task & $\ell$ & Raw ICON & CHOP(MFC) & Reduction \\
    \midrule
    \multirow{3}{*}{$\mathcal{M}_{g,\ell}^{1\to1}$: $\rho(0,x)\mapsto\rho(1,x)$}
      & 0.5 & $0.0281\pm0.0332$ & $0.0157\pm0.0087$ & 44.10\% \\
      & 0.3 & $0.2379\pm0.1401$ & $0.0364\pm0.0323$ & 84.69\% \\
      & 0.1 & $0.7627\pm0.0969$ & $0.1082\pm0.0686$ & 85.82\% \\
    \midrule
    \multirow{3}{*}{$\mathcal{M}_{g,\ell}^{1\to2}$: $\rho(0,x)\mapsto\rho_{[1/2,1]}(t,x)$}
      & 0.5 & $0.1097\pm0.0284$ & $0.0280\pm0.0187$ & 74.44\% \\
      & 0.3 & $0.1542\pm0.0453$ & $0.0334\pm0.0264$ & 78.37\% \\
      & 0.1 & $0.3219\pm0.0711$ & $0.0536\pm0.0358$ & 83.36\% \\
    \midrule
    \multirow{3}{*}{$\mathcal{M}_{g,\ell}^{2\to2}$: $\rho_{[0,1/2]}(t,x)\mapsto\rho_{[1/2,1]}(t,x)$}
      & 0.5 & $0.0314\pm0.0141$ & $0.0060\pm0.0034$ & 80.86\% \\
      & 0.3 & $0.1074\pm0.0375$ & $0.0156\pm0.0119$ & 85.47\% \\
      & 0.1 & $0.3095\pm0.0724$ & $0.0447\pm0.0308$ & 85.56\% \\
    \midrule
    \multirow{3}{*}{$\mathcal{M}_{\rho,\ell}^{1\to1}$: $g(x)\mapsto\rho(1,x)$}
      & 0.5 & $0.0046\pm0.0019$ & $0.0046\pm0.0019$ & 0.00\% \\
      & 0.3 & $0.0046\pm0.0018$ & $0.0046\pm0.0018$ & 0.00\% \\
      & 0.1 & $0.0040\pm0.0017$ & $0.0040\pm0.0017$ & 0.00\% \\
    \midrule
    \multirow{3}{*}{$\mathcal{M}_{\rho,\ell}^{1\to2}$: $g(x)\mapsto\rho_{[1/2,1]}(t,x)$}
      & 0.5 & $0.0795\pm0.0311$ & $0.0809\pm0.0356$ & $-1.69\%$ \\
      & 0.3 & $0.0747\pm0.0344$ & $0.0754\pm0.0366$ & $-0.97\%$ \\
      & 0.1 & $0.0700\pm0.0253$ & $0.0700\pm0.0253$ & 0.00\% \\
    \bottomrule
  \end{tabular}
\end{table}

We first ask whether a program evolved on
$\mathcal{M}_{g,0.5}^{2\to2}$ remains useful when the input-output geometry and
correlation length change while both sides of the mapping remain density fields.
Across these nine
density-to-density conditions, CHOP(MFC) reduces mean relative $L^2$ error by 44--86\%
(Fig.~\ref{fig:mfc_results}b; Table~\ref{tab:main}). Within each family, Raw ICON error and the
proportional reduction from CHOP both increase as $\ell$ decreases, with the
largest reductions at $\ell\in\{0.3,0.1\}$. In these mappings, the shared affine
statistics act on density values on both sides of the operator. The selected
examples in Fig.~\ref{fig:mfc_results}d illustrate the corresponding qualitative
pattern: Raw ICON smooths the sharp density peaks, while the CHOP predictions
follow them more closely.

Changing the mathematical role of the input produces a
different outcome. In the six cost-to-density conditions, the keys are
terminal-cost functions $g$ and the values are density fields, yet CHOP(MFC) still
pools their sampled values into one affine coordinate system. For
$\mathcal{M}_{\rho,\ell}^{1\to1}$, the in-context comparison selects Raw ICON for
every query, so the two error distributions and their arithmetic means coincide at
all three length scales. For $\mathcal{M}_{\rho,\ell}^{1\to2}$, mean error
increases by 1.69\% at $\ell=0.5$ and 0.97\% at $\ell=0.3$, and remains unchanged
at $\ell=0.1$. The complete program therefore lowers no condition-level mean
across these six tasks (Fig.~\ref{fig:mfc_results}c). This comparison alone cannot
determine whether the absence of improvement arises from the rescaled base
prediction, residual correction or the outer route decision, because all three
components act together in the complete program. This leaves a direct
component-level question: does the residual-correction module
$G_{\mathrm{corr}}$ remain useful when Raw ICON supplies its base predictions?
\subsection{Transfer of harness components}

The cost-to-density result motivates an ablation that isolates
the retained residual correction. We form CHOP(Corr) by placing
$G_{\mathrm{corr}}$ directly after Raw ICON. Raw ICON supplies the held-out and
complete-prompt base predictions, and $G_{\mathrm{corr}}$ applies the same
residual-transfer rule and internal gate selected during the original MFC
evolution. This reduced program omits the rescaled route and its outer comparison
with Raw ICON.

\begin{figure}[!htbp]
  \centering
  \includegraphics[width=0.995\textwidth]{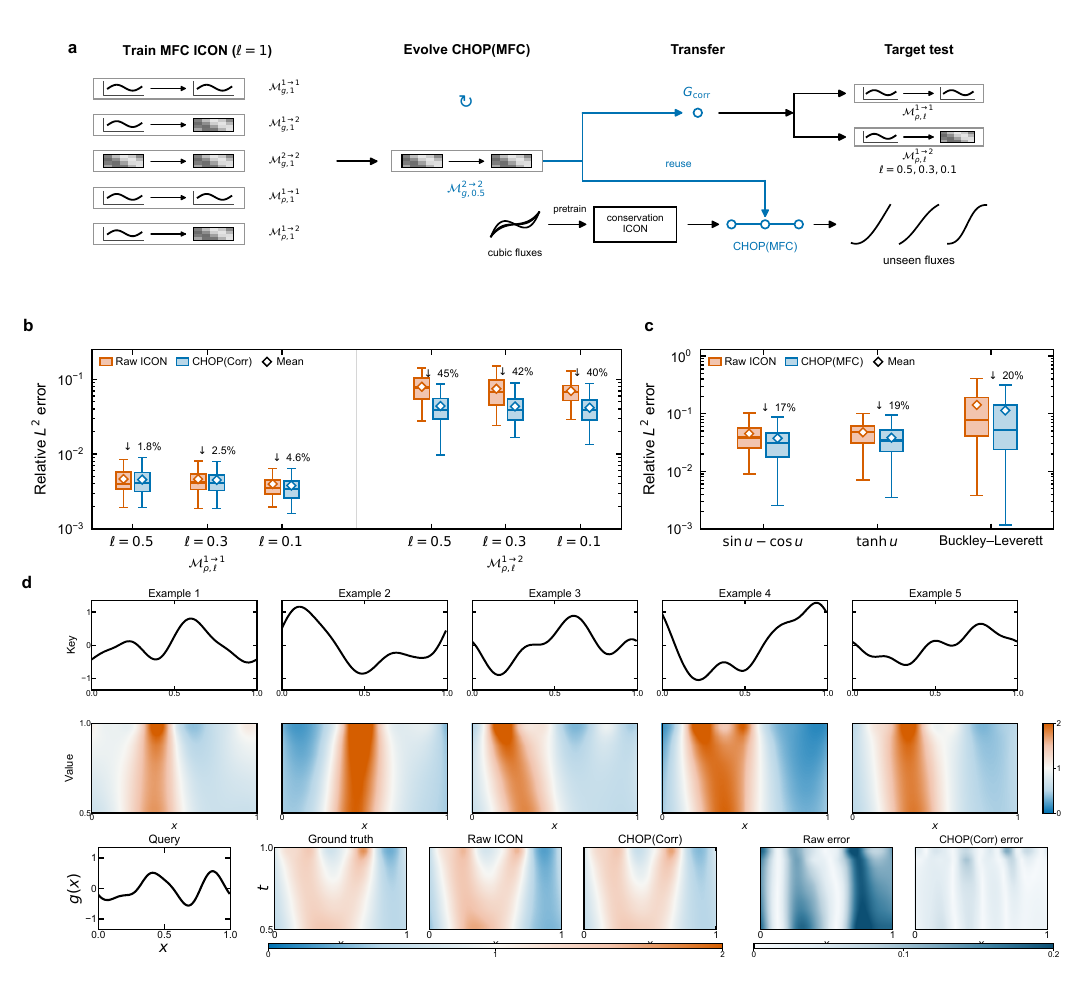}
  \caption{\textbf{CHOP components transfer across operator families and equation
  classes.}
  \textbf{a}, Transfer of the residual-correction operator
  $G_{\mathrm{corr}}$ across mean-field-control mappings and of the complete
  CHOP(MFC) program to conservation laws.
  \textbf{b}, Raw ICON and CHOP(Corr) relative $L^2$ error on cost-to-density
  mappings; each condition contains $n=100$ independently sampled operator
  instances.
  \textbf{c}, Raw ICON and transferred CHOP(MFC) relative $L^2$ error on
  non-cubic fluxes; each flux contains $n=500$ paired test trajectories.
  Percentages in \textbf{b} and \textbf{c} give reductions in the arithmetic
  mean from Raw ICON. Boxes span the interquartile range, centre lines mark
  medians, diamonds mark arithmetic means and whiskers extend to the most extreme
  values within $1.5\times$ the interquartile range; observations beyond the
  whiskers are not shown.
  \textbf{d}, One cost-to-density prompt, its predictions and pointwise absolute
  errors; the density and error colour scales are clipped at their displayed
  maxima.}
  \label{fig:operator_transfer}
\end{figure}

Without further evolution, we evaluate CHOP(Corr) on the same 100 prompts used
for each cost-to-density condition in Fig.~\ref{fig:mfc_results}c. It reduces
mean relative $L^2$ error by
1.8--4.6\% on $\mathcal{M}_{\rho,\ell}^{1\to1}$ and by 40--45\% on
$\mathcal{M}_{\rho,\ell}^{1\to2}$ (Fig.~\ref{fig:operator_transfer}b and
Supplementary Table~\ref{tab:mfc_component_ablation}).
Because Raw ICON and CHOP(Corr) are scored on the same original test prompts,
their paired comparison isolates the effect of the residual-correction module.
A comparison with the complete CHOP(MFC) program changes both the base route and
the outer decision, so it does not decompose the complete program's
cost-to-density outcome.

The two PDE settings provide a stricter test at the
level of the complete inference program. We couple CHOP(MFC), without further
evolution, to the conservation-law ICON introduced above. Every Raw ICON and
rescaled full-prompt or held-out evaluation now calls that frozen model, and the
resulting held-out errors drive the same residual correction and route decision
used in MFC.

On the same 500 trajectories for each non-cubic flux, the transferred program
reduces mean single-step relative $L^2$ error by 17--20\%
(Fig.~\ref{fig:operator_transfer}c). CHOP(Conservation) produces larger
reductions on the same trajectories (Table~\ref{tab:crosstask}).
The two programs differ in both their evolution tasks and their operator
structure, so this comparison establishes their performance ordering without
isolating one source of the difference. Together with the residual-only
ablation, the experiment demonstrates reuse at two levels: one correction
transfers across MFC operator mappings, and one complete program transfers
across frozen models and equation classes.

\begin{table}[!htbp]
  \centering
  \caption{\textbf{Cross-equation transfer of CHOP(MFC) to conservation-law fluxes.}
    Entries are arithmetic mean $\pm$ sample s.d.\ single-step relative $L^2$
    error over $n=500$ paired trajectories per flux; reductions are relative to
    the common Raw ICON mean.}
  \label{tab:crosstask}
  \small
  \setlength{\tabcolsep}{4.0pt}
  \begin{tabular}{lccc}
    \toprule
    Flux $f(u)$ & Raw ICON & CHOP(MFC) & CHOP(Conservation) \\
    \midrule
    $\sin u-\cos u$ & $0.0451\pm0.0312$ & $0.0374\pm0.0315$ & $0.0222\pm0.0219$ \\
    $\tanh u$           & $0.0475\pm0.0199$ & $0.0382\pm0.0212$ & $0.0313\pm0.0223$ \\
    $u^2/(u^2+(1-u)^2)$ & $0.1418\pm0.1530$ & $0.1138\pm0.1420$ & $0.0951\pm0.1307$ \\
    \bottomrule
  \end{tabular}
  \par\medskip
  \begin{tabular}{lcc}
    \toprule
    Flux $f(u)$ & CHOP(MFC) reduction & CHOP(Conservation) reduction \\
    \midrule
    $\sin u-\cos u$ & 17.06\% & 50.68\% \\
    $\tanh u$ & 19.48\% & 34.16\% \\
    $u^2/(u^2+(1-u)^2)$ & 19.78\% & 32.93\% \\
    \bottomrule
  \end{tabular}
\end{table}

\subsection{Real-world air-quality prediction}

\begin{figure}[!htbp]
  \centering
  \makebox[\textwidth][c]{%
    \includegraphics[width=\textwidth]{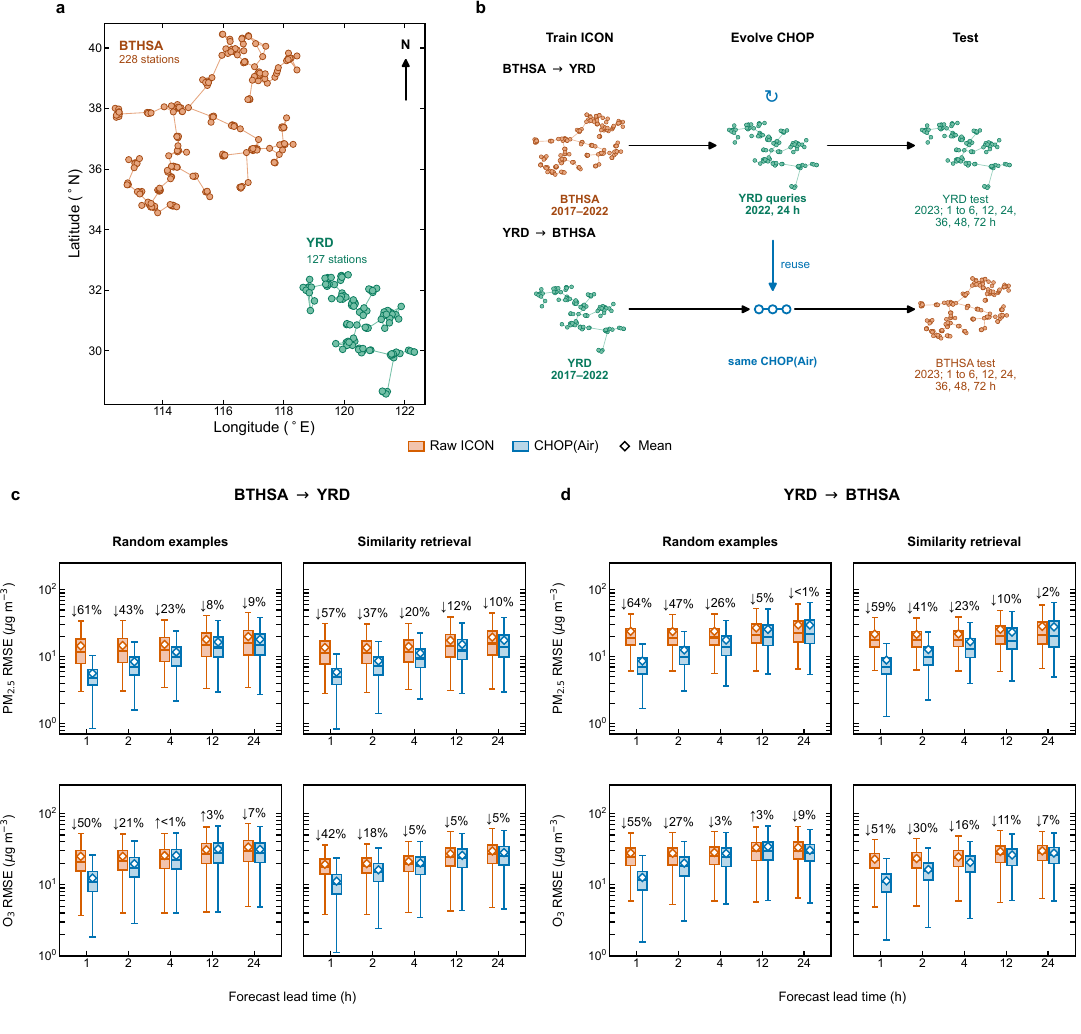}%
  }
  \caption{\textbf{Regional transfer setup and bidirectional air-quality
  prediction.}
  \textbf{a}, Monitoring-station graphs for the Beijing--Tianjin--Hebei and
  Surrounding Areas (BTHSA) and Yangtze River Delta (YRD). Displayed edges are
  subsampled; inference uses all edges in the $3^\circ$ station graphs.
  \textbf{b}, Forward evolution and reverse reuse of CHOP(Air) across the two
  regional ICONs.
  \textbf{c}, Per-case PM$_{2.5}$ and O$_3$ RMSE for BTHSA$\to$YRD with random
  and similarity-retrieved examples at the displayed forecast lead times.
  \textbf{d}, Corresponding YRD$\to$BTHSA results. Each observation is one
  target-time RMSE across all stations in the target region (127 for YRD and 228
  for BTHSA). At lead time $h$, each box contains $n=8{,}760-h$ valid 2023
  target times. Boxes span the interquartile range, centre lines mark medians,
  diamonds mark arithmetic means and whiskers extend to the most extreme values
  within $1.5\times$ the interquartile range; observations beyond the whiskers
  are not shown. Percentages give arithmetic-mean changes from Raw ICON;
  downward and upward arrows denote lower and higher CHOP(Air) error.}
  \label{fig:air}
\end{figure}

The controlled PDE experiments isolate particular forms of
distribution shift. We next ask whether inference-time orchestration of a frozen
model remains useful when the governing dynamics are unavailable and the source
and target regions have different station graphs. The frozen graph ICON predicts
the complete state of every station at a specified forecast lead time from a
24\,h window of meteorological variables, hour of day and pollutant
concentrations~\cite{wu2026graph}.
Each prompt
contains five example pairs and one query at the same lead: each example reveals
the future station state, while the query supplies only its input window.
Predictions use observations directly, without a chemical transport model.

We transfer between the Beijing--Tianjin--Hebei and Surrounding Areas (BTHSA)
graph, which contains 228 stations, and the Yangtze River Delta (YRD) graph,
which contains 127 stations (Fig.~\ref{fig:air}a). Examples are drawn from the
target-region historical record, either uniformly without replacement or by
similarity to the query window; every example target precedes the query origin.
All input channels enter the ICON evaluations and the distance used for the
nearest-example candidate.

The two directions assign distinct roles to the regions. For BTHSA$\to$YRD,
ICON is pretrained on BTHSA observations from 2017--2022 at leads sampled
uniformly from 1 to 24\,h. CHOP(Air) is evolved only in this direction on 200 YRD
24\,h prompts, split equally between the two example-selection strategies, with
examples from 2016--2021 and queries from 2022, and then fixed. Forward evaluation
uses 2023 YRD queries at 1--6, 12, 24, 36, 48 and 72\,h, with examples from
2016--2022; the last three leads exceed the pretraining range. For
YRD$\to$BTHSA, the same program is coupled without further evolution to ICON
pretrained on YRD observations from 2017--2022 and evaluated on 2023 BTHSA
queries with examples from 2016--2022 at the same leads.

CHOP(Air) uses in-context validation to combine seven pollutant
forecasts for each prompt.
Holding out each example output in turn evaluates seven candidate
routes against known targets. The candidates comprise Raw ICON, the first and last
input frames, the mean and nearest example outputs, and Raw ICON augmented by the
mean held-out residual, either directly or after graph-neighbor averaging. Their
errors fit
separate jackknife-stabilized simplex weights $a_{c,k}$ for PM$_{2.5}$ and O$_3$,
which are then applied to the corresponding candidates from the complete
prompt:
\begin{equation}
  \begin{aligned}
    \left(\{(u_i,v_i)\}_{i=1}^{D},u_*\right)
    &\xrightarrow{\mathrm{Raw\ ICON}} \widehat v_{*,1},\\
    \left(\{(u_i,v_i)\}_{i=1}^{D},u_*,\widehat v_{*,1}\right)
    &\xrightarrow{\mathrm{candidate\ construction}}
      \{\widehat v_{*,k}\}_{k=1}^{7},\\
    \widehat v_{*,c}
    &=\sum_{k=1}^{7}a_{c,k}\widehat v_{*,k,c},
      \qquad c\in\{\mathrm{PM}_{2.5},\mathrm O_3\}.
  \end{aligned}
\end{equation}
All other output channels retain their values from the
complete-prompt Raw ICON prediction $\widehat v_{*,1}$.
Supplementary Algorithm~\ref{alg:air_chain} specifies the exact
candidate construction, residual reuse and weight stabilization.

Similarity retrieval lowers Raw ICON mean per-case RMSE for both
pollutants, in both directions and at every evaluated lead time (Table~\ref{tab:air_forward_all_horizons} and
Supplementary Table~\ref{tab:air_reverse_all_horizons}). Comparing CHOP(Air) with this stronger
baseline tests whether the inference program adds further improvement after the
examples have been chosen for their relevance to the query.

\begin{table}[htbp]
  \centering
  \caption{\textbf{RMSE for BTHSA$\to$YRD.}
    At lead time $h$, entries are arithmetic means over $n=8{,}760-h$ valid 2023
    target times. Each per-case RMSE is computed across the 127 YRD stations and
    reported in $\mu$g m$^{-3}$. Reductions are relative to Raw ICON.}
  \label{tab:air_forward_all_horizons}
  \small
  \setlength{\tabcolsep}{3.2pt}
  \begin{tabular}{lrrrrrr}
    \toprule
    & \multicolumn{3}{c}{Random examples} & \multicolumn{3}{c}{Similarity retrieval} \\
    \cmidrule(lr){2-4}\cmidrule(lr){5-7}
    Lead time (h) & Raw & CHOP & Reduction (\%)
      & Raw & CHOP & Reduction (\%) \\
    \midrule
    \multicolumn{7}{l}{\textbf{PM$_{2.5}$}} \\
      1  & 14.56 &  5.63 & 61.3 & 13.73 &  5.88 & 57.2 \\
      2  & 14.70 &  8.43 & 42.6 & 13.70 &  8.66 & 36.8 \\
      3  & 14.94 & 10.37 & 30.6 & 13.89 & 10.24 & 26.3 \\
      4  & 15.25 & 11.76 & 22.8 & 14.22 & 11.33 & 20.3 \\
      5  & 15.59 & 12.86 & 17.5 & 14.63 & 12.16 & 16.9 \\
      6  & 15.96 & 13.70 & 14.2 & 15.07 & 12.84 & 14.8 \\
      12 & 18.08 & 16.55 &  8.4 & 17.53 & 15.36 & 12.4 \\
      24 & 19.96 & 18.20 &  8.8 & 19.62 & 17.67 &  9.9 \\
      36 & 21.27 & 19.59 &  7.9 & 20.91 & 19.49 &  6.8 \\
      48 & 22.34 & 20.02 & 10.4 & 21.97 & 20.29 &  7.7 \\
      72 & 23.79 & 20.93 & 12.0 & 23.22 & 21.22 &  8.6 \\
    \midrule
    \multicolumn{7}{l}{\textbf{O$_3$}} \\
      1  & 25.03 & 12.42 & 50.4 & 19.34 & 11.23 & 41.9 \\
      2  & 25.08 & 19.77 & 21.2 & 19.74 & 16.23 & 17.8 \\
      3  & 25.31 & 23.66 &  6.5 & 20.44 & 18.63 &  8.9 \\
      4  & 25.74 & 25.81 & $-0.3$ & 21.29 & 20.18 &  5.2 \\
      5  & 26.31 & 26.98 & $-2.6$ & 22.19 & 21.36 &  3.8 \\
      6  & 27.01 & 27.94 & $-3.4$ & 23.08 & 22.31 &  3.3 \\
      12 & 30.95 & 31.78 & $-2.7$ & 27.27 & 25.83 &  5.3 \\
      24 & 33.75 & 31.35 &  7.1 & 29.71 & 28.28 &  4.8 \\
      36 & 35.55 & 35.96 & $-1.2$ & 32.83 & 31.49 &  4.1 \\
      48 & 37.19 & 35.07 &  5.7 & 34.09 & 32.45 &  4.8 \\
      72 & 39.31 & 36.65 &  6.8 & 36.45 & 34.44 &  5.5 \\
    \bottomrule
  \end{tabular}
\end{table}

In the BTHSA$\to$YRD evaluation, CHOP(Air) improves on the stronger
similarity-retrieval baseline for both pollutants at all eleven lead times. At
1\,h, its reduction in mean per-case RMSE ranges from 42\% to 61\% across the
four pollutant and example-selection combinations. At 24\,h, all four means
remain lower by 5--10\%, showing that candidate combination adds to relevant
example selection even though the gains are smaller
(Fig.~\ref{fig:air}c and Table~\ref{tab:air_forward_all_horizons}). Representative 1\,h station fields
for one forward-transfer test day are shown in Supplementary
Fig.~\ref{fig:air_forecast_examples}.

The reverse test asks whether the program evolved only for
BTHSA$\to$YRD remains useful after both the frozen model and target station graph
change. Coupled without further evolution to the YRD-pretrained ICON, CHOP(Air)
reduces 1\,h mean per-case PM$_{2.5}$ RMSE by 64\% with random examples and 59\%
with similarity retrieval. The corresponding O$_3$ reductions are 55\% and 51\%
(Fig.~\ref{fig:air}d).

Across the complete sweep, CHOP(Air) lowers mean per-case RMSE
in 77 of the 88 pollutant, direction, example-selection and lead-time conditions.
The eleven increases range from 0.1\% to 3.4\% and are confined to reverse
PM$_{2.5}$ with similarity retrieval at 48 and 72\,h, and to O$_3$ with random
examples at five forward and four reverse lead times. Table~\ref{tab:air_forward_all_horizons} and
Supplementary Table~\ref{tab:air_reverse_all_horizons} report every lead time and the exact
magnitudes.

\FloatBarrier

\section{Discussion}

A standard
approach in scientific machine learning adapts a pretrained foundation model to
an out-of-distribution (OOD) scenario by adjusting its internal weights.
Our work with the Chain of Operators (CHOP) framework
studies a complementary route in which selected
physical and numerical distribution shifts are addressed at inference time
without updating a single model parameter. By treating a frozen foundation model
as a reusable computational primitive, CHOP shifts the focus from parameter tuning
to programmable inference. This approach instantiates a software-engineering
metaphor for scientific artificial intelligence: rather than rebuilding or
altering the neural engine, we construct a
mathematically specified scaffolding around it,
with explicit operations on
prompts and predictions.

 A major
advantage of the CHOP framework is
the auditability of its external adaptation logic.
The additional adaptation is governed by closed-form, explicit
elementary operators, such as coordinate scaling, moving frames, and residual
transfer. This architectural separation allows researchers to
inspect and test the added operations and
their dependencies. Furthermore, physical constraints, such as mass
conservation, can be enforced through an
exact output projection when the task justifies one. This modular design
makes the guarantee local and testable: the
conservation-law program enforces the stated mass constraint, while accuracy and
other physical properties remain empirical.

A particularly intriguing finding in our experiments is the transferability of
CHOP components
beyond their evolution task. Specifically,
the complete CHOP(MFC) program lowers mean error
when coupled without further evolution to a conservation-law ICON, while its
residual-correction module lowers mean error on the tested cost-to-density
mappings. This cross-task transferability
does not establish a universal error
structure, but it motivates studying a library of reusable, plug-and-play
inference operators that can be assembled dynamically to navigate new physical
systems.

Beyond its immediate performance gains, CHOP illustrates a structured division
of labor between human scientific expertise and machine intelligence. Human
expertise supplies the mathematical framework, physical constraints and
evaluation criteria, while agent search explores how these ingredients can be
assembled into executable programs. This division allows scientific ideas to
guide algorithm development without requiring every numerical choice to be
specified in advance. Joint interpretation and simplification give this
collaboration a further purpose: to develop the resulting programs into
numerical methods whose operations can be understood, tested and reused.

Despite these accomplishments, several open frontiers remain to be explored.
First, the selected
operator chains studied here are task-specific, meaning they are evolved offline
on a dedicated set and then fixed during evaluation.
Benefits vary, some programs require multiple ICON calls, and a
fresh search need not return identical source code; the fixed programs still
allow the recorded evaluation pipeline to regenerate the reported outputs and
metrics.
A natural and powerful extension of this paradigm is the development of an active
inference-time operator agent. For any newly encountered physical system, such an
agent could dynamically propose, compose, and evaluate candidate operator chains
on the fly through repeated probing calls to the frozen foundation model. Second,
the observed reuse across
operator mappings and PDE families warrants further mathematical investigation.
Future work should seek to characterize
when foundation-model errors support
reusable inference operators. Finally, while our transformations currently rely
on human-specified physics, developing methods that can automatically identify and
extract hidden coordinate invariances directly from a few contextual data examples
will allow CHOP to adapt to complex, completely black-box scientific systems where
the governing equations are entirely unknown.

\section*{Acknowledgements}
We acknowledge NUS IT's Research Computing group for providing computational
support.

\section*{Funding}
This work was supported by the National Research Foundation, Singapore, under
the NRF Fellowship (Project No. NRF-NRFF17-2025-0006) to L.Y.; and the National
Natural Science Foundation of China (grant nos. 92270115 and 12071301) to L.G.

\section*{Declaration of generative AI and AI-assisted technologies in the manuscript preparation process}
OpenAI Codex and Google Gemini assisted with manuscript writing and editing. The authors verified
all AI-assisted outputs and retain responsibility for the final manuscript,
programs, analyses and conclusions.

\section*{Data availability}
The air-quality observations used in this study are available in the KnowAir-V2
dataset at Zenodo~\cite{wang2025knowairv2}. Processed evaluation outputs for the
conservation-law, mean-field-control, cross-transfer and air-quality experiments
are publicly available at
\url{https://github.com/scaling-group/icon-chop}.

\section*{Code availability}
The implementations of the selected CHOP programs and the associated evaluation
code are publicly available at \url{https://github.com/scaling-group/icon-chop}.

\section*{Author contributions}
L.Y. conceptualized the study. M.Y. conducted the research. L.G. and L.Y.
supervised the work. M.Y., C.W., L.G. and L.Y. wrote and revised the manuscript.

\section*{Competing interests}
The authors declare no competing interests.

\appendix
\counterwithin{equation}{section}
\counterwithin{figure}{section}
\counterwithin{table}{section}
\counterwithin{algorithm}{section}
\section{Supplementary Methods}\label{sec:supp_methods}

Algorithms~\ref{alg:chop_G} and~\ref{alg:context_validated_combination} give the
particular residual-correction and candidate-combination rules used by the
selected programs in this work and are presented in the main Methods.
This appendix supplies numerical settings and the complete task-specific programs. Algorithm~\ref{alg:consv_chain_eval}
specifies the moving-frame transformation, shared rescaling and mass conservation
used by CHOP(Conservation). Algorithm~\ref{alg:chop_main} couples an
induced-operator route to Algorithm~\ref{alg:chop_G}, then selects between the
rescaled branch and Raw ICON. Algorithm~\ref{alg:air_chain} constructs the
candidate forecasts used by CHOP(Air) and combines them with
Algorithm~\ref{alg:context_validated_combination}.
Main-text Fig.~\ref{fig:induced_operator_route}
defines $F$, $G$ and $G^{-1}$ for the
abstract induced-operator route. The algorithms below identify each executable
stage by its action.

\subsection{Evolution and evaluation}
\label{sec:supp_chain_discovery}

We use the Evolving Ensemble of Agents framework~\cite{Eve} with OpenAI Codex as
its agent backend, accessed through a non-interactive Codex CLI, to search over
programs composed of explicit numerical operations. Researchers define the task,
operator library, physical constraints and evaluation objective; these choices
delimit the programs that Codex can propose. EvE orchestrates the Codex calls, and
a task evaluator scores executable candidates on the corresponding evolution set
while the ICON weights remain frozen.

The first candidates produced by EvE with Codex were executable, but their
complexity obscured the roles and dependencies of individual operations.
After the EvE search, researchers and Codex agents worked together to revise and
simplify these candidates until they could be inspected directly.
Every performance signal that guided this process came from the corresponding
evolution set. All revisions and selection decisions were completed before the
resulting program was run on any designated evaluation dataset. Because agent
proposals and human judgement both shaped this process, repeating it need not
produce identical CHOP source code. After the final program for a task was fixed,
it was run once on each designated evaluation dataset. Neither the evaluation
targets nor the resulting metrics were returned to the search or used for later
revision or tuning. The fixed program sources, evaluation pipeline, frozen model
checkpoints, evaluation data and recorded settings allow the reported evaluation
outputs and metrics to be regenerated.

For the PDE tasks, prediction error is measured using the relative $L^2$ error.
\begin{equation}
  \operatorname{Rel\text{-}}L^2(\widehat v,v)
  =\frac{\lVert \widehat v-v\rVert_2}{\lVert v\rVert_2}.
  \label{eq:supp_relative_l2}
\end{equation}
For air-quality prediction, we report the root-mean-square error for each
pollutant channel $c$,
\begin{equation}
  \operatorname{RMSE}_c(\widehat v,v)
  =\left[
    \frac{1}{N}\sum_{s=1}^{N}
    \bigl(\widehat v_c(s)-v_c(s)\bigr)^2
  \right]^{1/2},
  \label{eq:supp_air_rmse}
\end{equation}
where $N$ is the number of stations. Error reductions are reported relative
to Raw ICON evaluated on the same prompts.

\subsection{Numerical implementation of candidate combination}

The implementation applies projected gradient descent with exact Euclidean
projection onto $\Delta_K$ to every simplex problem in
Algorithm~\ref{alg:context_validated_combination}. For each quadratic problem,
let $\lambda_{\max}$ denote the largest
eigenvalue of its matrix. The implementation starts from the uniform vector,
uses step size $[2\max\{\lambda_{\max},10^{-12}\}]^{-1}$ and stops when the
largest coordinate change is at most $10^{-9}$ or after 512 iterations.

\subsection{Conservation-law experiments}
\label{sec:supp_conservation}

\subsubsection{Problem and experimental setting}

We consider the conservation law
\begin{equation}
  \partial_t u+\partial_x f(u)=0,
  \qquad x\in[0,1],
  \label{eq:supp_conservation_pde}
\end{equation}
with periodic boundary conditions. The frozen ICON is pretrained on cubic
fluxes $f(u)=au^3+bu^2+cu$, with $(a,b,c)$ sampled uniformly from
$[-1,1]^3$~\cite{yang2024pdegen}. The temporal step is $\Delta t=0.1$, and
initial conditions are sampled from a periodic Gaussian random field with
length scale $\ell=1$. Reference trajectories are computed with a
third-order weighted essentially non-oscillatory scheme~\cite{jiang1996efficient}.

Each numerical prompt contains $D=5$ one-step transitions from a single trajectory,
\begin{equation}
  (u_i,v_i)
  =\bigl(u(\cdot,t_i),u(\cdot,t_i+\Delta t)\bigr),
  \qquad i=1,\ldots,D,
  \label{eq:supp_conservation_prompt}
\end{equation}
where $t_i=t_1+(i-1)\Delta t$. The query is the last state in this history,
$u_*=u(\cdot,t_1+D\Delta t)$. During autoregressive
rollout, the five example transitions remain fixed and each predicted state
becomes the query for the next step.

CHOP(Conservation) is evolved on 100 trajectories generated with
$f(u)=\sin u-\cos u$, using single-step prediction error as fitness. The
selected chain is evaluated on 500 independent trajectories for each of
$f(u)=\sin u-\cos u$, $f(u)=\tanh u$ and
$f(u)=u^2/[u^2+(1-u)^2]$.

\subsubsection{CHOP(Conservation)}

The conservation-law prompt raises three numerical issues that can be resolved
before and after the ICON call. A traveling profile can make cyclic translation
dominate the observed one-step change, so the most recent example first
estimates a moving frame shared by the prompt. The shifted visible fields can
occupy different numerical ranges across prompts. The selected chain therefore
centers and rescales them with one prompt-derived affine transformation and
inverts that transformation after inference. The resulting numerical prediction
can violate the conserved mean,
so the final operator projects it onto the affine subspace with the query's mean.
Main-text equation~\eqref{eq:chop_conservation}
writes these transformations directly on
the example inputs, example outputs and query. For later reference, we abbreviate
the same intermediate fields and predictions in the chain
\begin{equation}
  \begin{aligned}
    \left(\{(u_i,v_i)\}_{i=1}^{D},u_*\right)
    &\xrightarrow{\mathrm{shift}}
    \left(\{(u_i^s,v_i^s)\}_{i=1}^{D},u_*^s\right)
    \\
    &\xrightarrow{\mathrm{rescale}}
    \left(\{(u_i^{\mathrm{sc}},v_i^{\mathrm{sc}})\}_{i=1}^{D},
      u_*^{\mathrm{sc}}\right)
    \xrightarrow{\mathrm{ICON}} \widehat v^{\mathrm{sc}},\\
    \widehat v^{\mathrm{sc}}
    &\xrightarrow{\mathrm{inverse\ rescale}} \widehat v_s
    \xrightarrow{\mathrm{inverse\ shift}} \widehat v
    \xrightarrow{\mathrm{mass\ conservation}} \widehat v_*.
  \end{aligned}
  \label{eq:supp_conservation_chain}
\end{equation}
For a field $v$ sampled on $N_x$ spatial grid points, let
$\langle v\rangle$ denote its spatial mean and
$\operatorname{shift}(v,r)$ its cyclic shift by $r$ grid points. The most recent
example pair $(u_D,v_D)$ estimates the spatial shift $s$. The forward shift
removes this motion from the numerical prompt by shifting a field at temporal
index $n$ by $-ns$.
The temporal indices of $u_i$, $v_i$ and $u_*$ are $i-1$, $i$ and $D$,
respectively, giving the shifts $-(i-1)s$, $-is$ and $-Ds$. Rescaling pools all
grid points from the shifted numerical prompt. Their mean and standard deviation
define the center
$m_{\mathrm{sc}}$ and scale $\sigma_{\mathrm{sc}}$, respectively,
\begin{equation}
  \begin{aligned}
    m_{\mathrm{sc}}
    &=\operatorname{mean}(u_1^s,v_1^s,\ldots,u_D^s,v_D^s,u_*^s),\\
    \sigma_{\mathrm{sc}}
    &=\max\!\left\{
      \operatorname{std}(u_1^s,v_1^s,\ldots,u_D^s,v_D^s,u_*^s),
      \varepsilon_{\mathrm{mach}}\right\}.
  \end{aligned}
  \label{eq:supp_conservation_rescaling_statistics}
\end{equation}
$\operatorname{std}$ in equation~\eqref{eq:supp_conservation_rescaling_statistics}
is the sample standard deviation, and $\varepsilon_{\mathrm{mach}}$ denotes
machine precision for the tensor data type used in the computation.
Each field $u$ in the shifted numerical prompt becomes
$u^{\mathrm{sc}}=(u-m_{\mathrm{sc}})/\sigma_{\mathrm{sc}}$.
After ICON inference, the inverse rescaling restores the value scale. The
prediction has temporal index $D+1$ and is therefore shifted by $(D+1)s$ to
restore its original spatial position.
Finally, the mass-conservation step enforces the conserved mean.

\begin{algorithm}[!ht]
\caption{CHOP(Conservation).}
\label{alg:consv_chain_eval}
\begin{algorithmic}[1]
\small
\Require $D=5$ pairs $\{(u_i,v_i)\}_{i=1}^{D}$ sampled on $N_x\geq4$ spatial
  grid points, query $u_*$ and frozen ICON.
\Output Prediction $\widehat v_*$.

\OperatorComment{Shift: estimate the displacement and enter the moving frame}
\State $\bar u_D\leftarrow u_D-\langle u_D\rangle$;
  $\bar v_D\leftarrow v_D-\langle v_D\rangle$.
\State $s\leftarrow
  \underset{|r|\leq\lfloor N_x/4\rfloor,\ r\in\mathbb Z}{\arg\min}
  \lVert\operatorname{shift}(\bar u_D,r)-\bar v_D\rVert_2^2$.
\State If the minimum is attained by several shifts, choose the smallest $r$.
\For{$i=1,\ldots,D$}
  \State $u_i^s\leftarrow\operatorname{shift}(u_i,-(i-1)s)$;
    $v_i^s\leftarrow\operatorname{shift}(v_i,-is)$.
\EndFor
\State $u_*^s\leftarrow\operatorname{shift}(u_*,-Ds)$.

\OperatorComment{Rescale: apply one shared affine transformation}
\State Compute $m_{\mathrm{sc}}$ and set $\sigma_{\mathrm{sc}}$ to the larger
  of the pooled standard deviation and $\varepsilon_{\mathrm{mach}}$ over all grid points in
  $\{u_i^s,v_i^s\}_{i=1}^{D}\cup\{u_*^s\}$.
\State Set
  $u_i^{\mathrm{sc}}\leftarrow(u_i^s-m_{\mathrm{sc}})/\sigma_{\mathrm{sc}}$,
  $v_i^{\mathrm{sc}}\leftarrow(v_i^s-m_{\mathrm{sc}})/\sigma_{\mathrm{sc}}$
  and
  $u_*^{\mathrm{sc}}\leftarrow(u_*^s-m_{\mathrm{sc}})/\sigma_{\mathrm{sc}}$.

\OperatorComment{ICON: predict from the shifted, rescaled prompt}
\State $\widehat v^{\mathrm{sc}}\leftarrow
  \mathrm{ICON}(\{(u_i^{\mathrm{sc}},v_i^{\mathrm{sc}})\}_{i=1}^{D},
  u_*^{\mathrm{sc}})$.
\OperatorComment{Inverse rescale: restore the original value scale}
\State $\widehat v_s\leftarrow
  \sigma_{\mathrm{sc}}\widehat v^{\mathrm{sc}}+m_{\mathrm{sc}}$.
\OperatorComment{Inverse shift: restore the original spatial frame}
\State $\widehat v\leftarrow\operatorname{shift}(\widehat v_s,(D+1)s)$.

\OperatorComment{Mass conservation: enforce the conserved mean}
\State \Return $\widehat v_*\leftarrow
  \widehat v+\langle u_*\rangle-\langle\widehat v\rangle$.
\end{algorithmic}
\end{algorithm}

\subsection{Mean-field control experiments}
\label{sec:supp_mfc}

\subsubsection{Problem and experimental setting}

The mean-field control problem minimizes the transport and terminal costs
\begin{equation}
  \inf_{\rho,m}
  \int_0^1\!\!\int_0^1
  \frac{c\,m^2(t,x)}{2\rho(t,x)}\,dx\,dt
  +\int_0^1 g(x)\rho(1,x)\,dx,
  \label{eq:supp_mfc_objective}
\end{equation}
subject to
\begin{equation}
  \partial_t\rho+\partial_xm=\mu\,\partial_{xx}\rho,
  \label{eq:supp_mfc_constraint}
\end{equation}
with periodic boundary conditions, transport-cost coefficient $c=20$ and
viscosity $\mu=0.02$. The frozen ICON is pretrained on five operator
families~\cite{liu2023incontext}, which comprise
$\mathcal M_{g,\ell}^{1\to1}:\rho(0,x)\mapsto\rho(1,x)$,
$\mathcal M_{g,\ell}^{1\to2}:\rho(0,x)\mapsto\rho_{[1/2,1]}(t,x)$,
$\mathcal M_{g,\ell}^{2\to2}:\rho_{[0,1/2]}(t,x)\mapsto\rho_{[1/2,1]}(t,x)$,
$\mathcal M_{\rho,\ell}^{1\to1}:g(x)\mapsto\rho(1,x)$ and
$\mathcal M_{\rho,\ell}^{1\to2}:g(x)\mapsto\rho_{[1/2,1]}(t,x)$. The
superscript records the dimensions of the input and output domains, and
$\rho_{[a,b]}$ restricts the density trajectory to $t\in[a,b]$. Within each
$\mathcal M_{g,\ell}$ prompt, one sampled terminal cost $g$ is fixed, while the
examples and query vary the initial density directly or the early-time trajectory
generated from it, according to the mapping. Within each $\mathcal M_{\rho,\ell}$
prompt, one sampled initial density is fixed, while the examples and query vary
$g$. The function $g(x)$ or $\rho(0,x)$ is sampled from a Gaussian process with
a radial basis function kernel and length scale $\ell=1$. Test length scales
$\ell\in\{0.5,0.3,0.1\}$ generate 15 out-of-distribution tasks.
Every prompt contains $D=5$ example pairs and one query. CHOP(MFC) is evolved
using 100 independent operator instances from
$\mathcal M_{g,0.5}^{2\to2}:\rho_{[0,1/2]}\mapsto\rho_{[1/2,1]}$. The other 14
tasks are excluded from evolution, and final evaluation uses 100 independent
instances for each of the 15 tasks.

\subsubsection{CHOP(MFC)}
\label{sec:supp_chop_mfc}

CHOP(MFC) makes two nested decisions from errors observed at the example
inputs. Within the rescaled branch, Algorithm~\ref{alg:chop_G} fits
$\hat\alpha$ and applies the residual correction when
$E_{\mathrm{fit}}$ is finite and $E_{\mathrm{fit}}<E_{\mathrm{base}}$. The
outer decision compares the same
pre-gate value $E_{\mathrm{fit}}$ with the Raw ICON error
$E_{\mathrm{raw}}$. Consequently, the outer decision can select the rescaled
branch after the internal gate has set $\alpha_{\mathrm{use}}=0$; that branch
then returns its uncorrected base prediction.
The rescaled base route rescales the prompt, evaluates ICON and
inverts the rescaling in sequence:
\begin{equation}
  \left(\{(u_i,v_i)\}_{i=1}^{D},u\right)
  \xrightarrow{\mathrm{rescale}}
  \left(\{(u_i^{\mathrm{sc}},v_i^{\mathrm{sc}})\}_{i=1}^{D},
    u^{\mathrm{sc}}\right)
  \xrightarrow{\mathrm{ICON}}\widehat v^{\mathrm{sc}}
  \xrightarrow{\mathrm{inverse\ rescale}}\widehat v_{\mathrm{base}}(u).
  \label{eq:supp_mfc_rescaling_chain}
\end{equation}

Before each ICON evaluation on this route, CHOP(MFC) applies a shared
rescaling to the numerical prompt. Let $z_{\mathrm{vis}}$ be the vector formed by
concatenating every sampled value in the example inputs, example outputs and
query input. The transformation center $m_{\mathrm{sc}}$ and scale
$\sigma_{\mathrm{sc}}$ are computed from this vector. When
$\min z_{\mathrm{vis}}\geq-\varepsilon_{\mathrm{mach}}$, the code uses
\begin{equation}
  m_{\mathrm{sc}}=0,
  \qquad
  \sigma_{\mathrm{sc}}
  =\sqrt{\max\!\left\{\operatorname{mean}(z_{\mathrm{vis}}^2),
                       \varepsilon_{\mathrm{mach}}\right\}},
  \label{eq:supp_mfc_nonnegative_rescaling}
\end{equation}
which preserves non-negativity. For sign-changing prompts,
$m_{\mathrm{sc}}=\operatorname{mean}(z_{\mathrm{vis}})$ and
$\sigma_{\mathrm{sc}}
=\max\{\operatorname{std}(z_{\mathrm{vis}}),\varepsilon_{\mathrm{mach}}\}$, where
$\operatorname{std}$ is the sample standard deviation. The same transformation
is applied to every field,
\begin{equation}
  u_i^{\mathrm{sc}}=\frac{u_i-m_{\mathrm{sc}}}{\sigma_{\mathrm{sc}}},
  \qquad
  v_i^{\mathrm{sc}}=\frac{v_i-m_{\mathrm{sc}}}{\sigma_{\mathrm{sc}}},
  \qquad
  u_*^{\mathrm{sc}}=\frac{u_*-m_{\mathrm{sc}}}{\sigma_{\mathrm{sc}}},
  \label{eq:supp_mfc_rescaling}
\end{equation}
and the base prediction is returned to the original scale as
\begin{equation}
  \widehat v_{*,\mathrm{base}}
  =\sigma_{\mathrm{sc}}
    \mathrm{ICON}(\{(u_i^{\mathrm{sc}},v_i^{\mathrm{sc}})\}_{i=1}^{D},
      u_*^{\mathrm{sc}})
   +m_{\mathrm{sc}}.
  \label{eq:supp_mfc_rescaled_prediction}
\end{equation}

CHOP(MFC) supplies this prediction to Algorithm~\ref{alg:chop_G} together with
$s_u=\sigma_{\mathrm{sc}}$. For each example $h$, the code removes pair $h$,
preserves the order of the remaining pairs and appends the first remaining pair
so that ICON still receives $D$ examples, with $D=5$ in every experiment. The
rescaling statistics are recomputed on this augmented prompt. Applying the same
three-component route at $u_h$ gives $\widehat v_{h,\mathrm{base}}^{(-h)}$ and
the observable residual
\begin{equation}
  R_h=v_h-\widehat v_{h,\mathrm{base}}^{(-h)}.
  \label{eq:supp_mfc_loo_residual}
\end{equation}
With these inputs, Algorithm~\ref{alg:chop_G} returns the corrected branch
$\widehat v_{*,\mathrm{corr}}$ and its fitted residual error
$E_{\mathrm{fit}}$.

Raw ICON is evaluated independently on the same augmented prompts. Its
prediction at $u_h$ gives the residual
\begin{equation}
  R_{h,\mathrm{raw}}
  =v_h-\widehat v_{h,\mathrm{raw}}^{(-h)},
  \label{eq:supp_mfc_raw_loo_residual}
\end{equation}
and evaluating Raw ICON on the complete prompt gives
\begin{equation}
  \widehat v_{*,\mathrm{raw}}
  =\mathrm{ICON}\!\left(\{(u_i,v_i)\}_{i=1}^{D},u_*\right).
  \label{eq:supp_mfc_raw_query_prediction}
\end{equation}
The Raw ICON error visible at the example inputs is
\begin{equation}
  E_{\mathrm{raw}}
  =\frac{1}{D}\sum_{h=1}^{D}
    \operatorname{mean}(R_{h,\mathrm{raw}}^2).
  \label{eq:supp_mfc_raw_error}
\end{equation}
The pre-gate $E_{\mathrm{fit}}$ returned by Algorithm~\ref{alg:chop_G} and the
Raw ICON error determine the final query route:
\begin{equation}
  \widehat v_{*,\mathrm{MFC}}
  =\begin{cases}
    \widehat v_{*,\mathrm{corr}},
      & E_{\mathrm{fit}}\text{ finite and }E_{\mathrm{fit}}<E_{\mathrm{raw}},\\
    \widehat v_{*,\mathrm{raw}}, & \text{otherwise}.
  \end{cases}
  \label{eq:supp_mfc_corrected_prediction}
\end{equation}

For $D=5$, the two sets of held-out evaluations and the two complete-prompt
evaluations give $Q=2D+2=12$ logical ICON evaluations per query.

\begin{algorithm}[!ht]
\caption{CHOP(MFC).}
\label{alg:chop_main}
\begin{algorithmic}[1]
\small
\Require $D\geq2$ pairs $\{(u_i,v_i)\}_{i=1}^{D}$, query $u_*$ and
  frozen ICON.
\Output Prediction $\widehat v_{*,\mathrm{MFC}}$.

\OperatorComment{In-context validation: construct the held-out prompts}
\For{$h=1,\ldots,D$}
  \State Remove pair $h$ while preserving order, then append the first
    remaining pair to form an augmented prompt.
\EndFor
\OperatorComment{Rescaled base route: predict the held-out outputs}
\State Evaluate the rescaled route on the $D$ augmented prompts to obtain
  $\{\widehat v_{h,\mathrm{base}}^{(-h)}\}_{h=1}^{D}$, recomputing the
  rescaling statistics for every prompt.
\OperatorComment{Raw ICON: predict the held-out outputs from the prompts as given}
\State Evaluate Raw ICON on the same augmented prompts to obtain
  $\{\widehat v_{h,\mathrm{raw}}^{(-h)}\}_{h=1}^{D}$.
\OperatorComment{In-context validation: compute the Raw ICON error}
\State Compute $\{R_{h,\mathrm{raw}}\}_{h=1}^{D}$ from
  equation~\eqref{eq:supp_mfc_raw_loo_residual}.
\State Compute $E_{\mathrm{raw}}$ from equation~\eqref{eq:supp_mfc_raw_error}.

\OperatorComment{Rescaled base route: predict from the complete prompt}
\State Compute $(m_{\mathrm{sc}},\sigma_{\mathrm{sc}})$ from the complete
  prompt using equation~\eqref{eq:supp_mfc_nonnegative_rescaling} and the
  sign-changing rule above.
\State Evaluate the rescaled route to obtain
  $\widehat v_{*,\mathrm{base}}$ from
  equation~\eqref{eq:supp_mfc_rescaled_prediction}.
\OperatorComment{Raw ICON: predict from the complete prompt as given}
\State Evaluate Raw ICON on the complete prompt to obtain
  $\widehat v_{*,\mathrm{raw}}$ from
  equation~\eqref{eq:supp_mfc_raw_query_prediction}.

\OperatorComment{Residual correction: fit its magnitude and apply the internal gate}
\State Apply Algorithm~\ref{alg:chop_G} to the example pairs and query with
  $\{\widehat v_{h,\mathrm{base}}^{(-h)}\}_{h=1}^{D}$,
  $\widehat v_{*,\mathrm{base}}$ and $s_u\leftarrow\sigma_{\mathrm{sc}}$ to obtain
  $\widehat v_{*,\mathrm{corr}}$ and $E_{\mathrm{fit}}$.
\OperatorComment{In-context validation: select the rescaled branch or Raw ICON}
\State Set $\widehat v_{*,\mathrm{MFC}}$ from
  equation~\eqref{eq:supp_mfc_corrected_prediction}.
\State Reject $\widehat v_{*,\mathrm{MFC}}$ if any entry is non-finite.
\State \Return $\widehat v_{*,\mathrm{MFC}}$.
\end{algorithmic}
\end{algorithm}

\subsection{Air-quality experiments}
\label{sec:supp_air}

\subsubsection{Problem and experimental setting}

The air-quality model performs in-context operator inference by message
passing on a graph whose nodes are monitoring stations~\cite{wu2026graph}. A
model trained on one region is transferred to the other region without weight
updates. Each source-region model is trained on data from 2017--2022
with forecast lead times drawn uniformly from 1 to 24\,h. For
geometric generalization, example pairs are retrieved from data collected
in the other region between 2016 and 2022. Across the two regions, the dataset
contains 70,128 hourly time points from 2016 through 2023.

Each numerical prompt contains $D=5$ example pairs and one query. An example
input and the query are 24\,h windows of station observations, and their outputs
are station fields at the specified lead time. Example pairs are selected either
uniformly without replacement from eligible historical windows or from a
precomputed similarity index for the query. In both cases, every example target
precedes the query.

CHOP(Air) is evolved only for the BTHSA-to-YRD transfer direction. Evolution
uses the BTHSA-trained model and 200 YRD prompts at the 24\,h lead time, split
equally between random and similarity-based example retrieval. Example pairs
are restricted to 2016--2021 and evolution queries are sampled
from 2022. The resulting program is fixed before evaluation. Final evaluation
uses data from 2023 at lead times of 1 through 6, 12, 24, 36, 48 and 72\,h in
both geographic-transfer directions, with examples retrieved from 2016--2022.
The first eight lead times lie within the range used during pretraining. The
remaining three evaluate direct forecasts beyond that range.

\subsubsection{Candidate predictions}
\label{sec:supp_air_candidates}

Each predicted station state contains the meteorological and hour-of-day
channels first, followed by PM$_{2.5}$ and O$_3$. CHOP(Air) combines seven
candidates for these final two pollutant channels; the preceding channels
retain the Raw ICON prediction. It first
evaluates Raw ICON at each example input using the remaining $D-1$ pairs. For
example $h$, the prediction and its observable residual are
\begin{equation}
  \widehat v_{h,\mathrm{raw}}^{(-h)}
  =\operatorname{ICON}\!\left(\{(u_j,v_j)\mid j\ne h\},u_h\right),
  \qquad
  R_{h,\mathrm{raw}}=v_h-\widehat v_{h,\mathrm{raw}}^{(-h)}.
  \label{eq:supp_air_loo_residual}
\end{equation}
The $D$ residuals $\{R_{h,\mathrm{raw}}\}_{h=1}^{D}$ are computed once and
retained for candidate construction.
Each Air held-out prompt contains exactly the remaining $D-1$ pairs.

The seven candidates draw on four sources. Raw ICON supplies the model
prediction. One candidate copies the last input frame and another copies the
first. The mean and nearest example outputs supply two
alternatives from the observed pairs. The remaining candidates add the mean
stored ICON residual to Raw ICON, either directly or after averaging that
residual over neighboring stations.

\begingroup\sloppy
Write a chronologically ordered 24-frame input window as
$u=(u^{(1)},\ldots,u^{(24)})$, where $u^{(t)}$ is the complete station field
in frame $t$. The first- and last-frame projections are\par\endgroup
\begin{equation}
  p_{\mathrm{first}}(u)=u^{(1)},
  \qquad
  p_{\mathrm{last}}(u)=u^{(24)}.
  \label{eq:supp_air_endpoint_frames}
\end{equation}
To construct the candidates at example input $u_h$, define
\begin{equation}
  \bar v_{-h}=\frac{1}{D-1}\sum_{\substack{j=1\\j\ne h}}^{D}v_j,
  \qquad
  \bar R_{-h}=\frac{1}{D-1}\sum_{\substack{j=1\\j\ne h}}^{D}R_{j,\mathrm{raw}}.
  \label{eq:supp_air_holdout_means}
\end{equation}
Every remaining residual is reused from
equation~\eqref{eq:supp_air_loo_residual}; for $j\ne h$, the ICON prompt used to
compute $R_{j,\mathrm{raw}}$ contains pair $h$. Thus, $\bar R_{-h}$ excludes
$R_{h,\mathrm{raw}}$ but can depend on $v_h$ through the retained ICON
predictions. Candidates 6 and 7 inherit this dependence. The combination module
receives each candidate row exactly as constructed.
The nearest example is determined from the distance
\begin{equation}
  d_{hj}
  =\operatorname{mean}_{t,s,q}\!\left[
    \left(
      \frac{u_{j,t,s,q}-u_{h,t,s,q}}{\sigma_q^{\mathrm{val}}}
    \right)^2
  \right],
  \qquad
  j_h=\underset{j\ne h}{\arg\min}\ d_{hj},
  \label{eq:supp_air_holdout_distance}
\end{equation}
where $t$, $s$ and $q$ index time, station and input feature. The feature
scale is
$\sigma_q^{\mathrm{val}}=\max\{\operatorname{std}_{i,t,s}
(u_{i,t,s,q}),10^{-4}\}$, with $i=1,\ldots,D$ and
$\operatorname{std}$ denoting the population standard deviation. If several
examples attain the same minimum distance, the implementation selects the
first one in the preserved prompt order. This rule applies to $j_h$ and to the
query index $j_*$ defined below.

The graph operator $\mathcal A_G$ averages each station residual over its
neighbors. Let $\mathcal N_G(s)$ denote the neighbors of station $s$. For a
station field $z$,
\begin{equation}
  [\mathcal A_G(z)]_{s,c}
  =\begin{cases}
    |\mathcal N_G(s)|^{-1}
      \displaystyle\sum_{r\in\mathcal N_G(s)}z_{r,c},
      &|\mathcal N_G(s)|>0,\\[4pt]
    z_{s,c},&|\mathcal N_G(s)|=0.
  \end{cases}
  \label{eq:supp_air_graph_average}
\end{equation}
Edges are treated as undirected and their weights are ignored. An isolated
station therefore retains its own residual.

The seven candidate predictions at example input $u_h$ are
\begin{equation}
  \begin{aligned}
    \widehat v_{h,1}&=\widehat v_{h,\mathrm{raw}}^{(-h)},
    &\widehat v_{h,2}&=p_{\mathrm{last}}(u_h),
    &\widehat v_{h,3}&=p_{\mathrm{first}}(u_h),\\
    \widehat v_{h,4}&=\bar v_{-h},
    &\widehat v_{h,5}&=v_{j_h},
    &\widehat v_{h,6}&=\widehat v_{h,\mathrm{raw}}^{(-h)}+\bar R_{-h},\\
    \widehat v_{h,7}&=\widehat v_{h,\mathrm{raw}}^{(-h)}+\mathcal A_G(\bar R_{-h}).
  \end{aligned} 
  \label{eq:supp_air_holdout_candidates}
\end{equation}

For the query $u_*$, let
\begin{equation}
  \bar v=\frac{1}{D}\sum_{h=1}^{D}v_h,
  \qquad
  \bar R=\frac{1}{D}\sum_{h=1}^{D}R_{h,\mathrm{raw}}.
  \label{eq:supp_air_query_means}
\end{equation}
Let $j_*$ minimize the distance between $u_*$ and the example inputs,
computed as in equation~\eqref{eq:supp_air_holdout_distance} after recomputing
the feature scale from the $D$ example inputs together with $u_*$. The
candidate predictions for the query are
\begin{equation}
  \begin{aligned}
    \widehat v_{*,1}&=\operatorname{ICON}\!\left(
      \{(u_i,v_i)\}_{i=1}^{D},u_*\right),
    &\widehat v_{*,2}&=p_{\mathrm{last}}(u_*),
    &\widehat v_{*,3}&=p_{\mathrm{first}}(u_*),\\
    \widehat v_{*,4}&=\bar v,
    &\widehat v_{*,5}&=v_{j_*},
    &\widehat v_{*,6}&=\widehat v_{*,1}+\bar R,\\
    \widehat v_{*,7}&=\widehat v_{*,1}+\mathcal A_G(\bar R).
  \end{aligned}
  \label{eq:supp_air_query_candidates}
\end{equation}

\begin{algorithm}[!t]
\caption{CHOP(Air).}
\label{alg:air_chain}
\begin{algorithmic}[1]
\small
\Require $D\geq2$ pairs $\{(u_i,v_i)\}_{i=1}^{D}$, query $u_*$,
  frozen ICON and target-region graph $G$.
\Output Prediction $\widehat v_*$.

\OperatorComment{In-context validation: construct the held-out prompts}
\State Remove each pair $h$ in turn while preserving the order of the remaining
  $D-1$ pairs, and use $u_h$ as the corresponding query.
\OperatorComment{Raw ICON: predict the held-out outputs}
\State Evaluate Raw ICON on the $D$ held-out prompts to obtain
  $\{\widehat v_{h,\mathrm{raw}}^{(-h)}\}_{h=1}^{D}$.
\State Replace any NaN or infinite entries in these predictions by zero.
\OperatorComment{In-context validation: compute the observed residuals}
\State Compute $\{R_{h,\mathrm{raw}}\}_{h=1}^{D}$ from
  equation~\eqref{eq:supp_air_loo_residual}.

\OperatorComment{Candidate construction: assemble one row at each example input}
\For{$h=1,\ldots,D$}
  \State Generate $\{\widehat v_{h,k}\}_{k=1}^{7}$ using
    equations~\eqref{eq:supp_air_endpoint_frames}--\eqref{eq:supp_air_holdout_candidates}.
\EndFor

\OperatorComment{Candidate-weight fitting: fit and stabilize pollutant-specific weights}
\For{$c\in\{\mathrm{PM}_{2.5},\mathrm O_3\}$}
  \State Compute $\{M_{h,c}\}_{h=1}^{D}$ and $M_c$ using
    equation~\eqref{eq:supp_air_error_matrix}.
  \State Using $M_c$, compute $a_c$ from
    equations~\eqref{eq:candidate_weight_fit}--\eqref{eq:candidate_weight_shrinkage}
    with $K=7$ and $a_{\mathrm{ref}}=(1,0,\ldots,0)^\top$.
\EndFor

\OperatorComment{Query candidates: evaluate Raw ICON and construct the others}
\State Recompute $\sigma_q^{\mathrm{val}}$ from the $D$ example inputs and
  $u_*$, then choose the nearest example $j_*$ using
  equation~\eqref{eq:supp_air_holdout_distance}.
\State Evaluate Raw ICON on the complete prompt to obtain $\widehat v_{*,1}$,
  then construct $\{\widehat v_{*,k}\}_{k=2}^{7}$ using
  equation~\eqref{eq:supp_air_endpoint_frames} and
  equations~\eqref{eq:supp_air_query_means}--\eqref{eq:supp_air_query_candidates}.
\OperatorComment{Candidate combination: apply the pollutant-specific weights}
\State Compute $\widehat v_{*,c}$ from
  equation~\eqref{eq:supp_air_channel_combination} for
  $c\in\{\mathrm{PM}_{2.5},\mathrm O_3\}$.
\State Set the pollutant channels of $\widehat v_*$ to
  $\{\widehat v_{*,c}\}$ and retain the preceding channels from
  $\widehat v_{*,1}$.
\State Replace any NaN or infinite entries in $\widehat v_*$ by zero.
\State \Return $\widehat v_*$.
\end{algorithmic}
\end{algorithm}

\subsubsection{Pollutant-specific candidate combination}
\label{sec:supp_air_combination}

The supplied candidate rows instantiate
Algorithm~\ref{alg:context_validated_combination} with $K=7$ separately for
PM$_{2.5}$ and O$_3$. Candidate 1 is Raw ICON, so the reference vector
$a_{\mathrm{ref}}=(1,0,\ldots,0)^{\top}$ selects it. For
$c\in\{\mathrm{PM}_{2.5},\mathrm{O}_3\}$, let
$a_c=(a_{c,1},\ldots,a_{c,7})^{\top}$ denote the candidate weights. The error
matrix at example input $u_h$ is
\begin{equation}
  \begin{aligned}
    (M_{h,c})_{kr}
    &=\frac{1}{N}\sum_{s=1}^{N}
      \left(\widehat v_{h,k,c}(s)-v_{h,c}(s)\right)
      \left(\widehat v_{h,r,c}(s)-v_{h,c}(s)\right),\\
    M_c&=\frac{1}{D}\sum_{h=1}^{D}M_{h,c}.
  \end{aligned}
  \label{eq:supp_air_error_matrix}
\end{equation}
For each pollutant, Algorithm~\ref{alg:context_validated_combination} uses
$M=M_c$. Each delete-one refit omits one already constructed matrix $M_{h,c}$
from the average and reuses the remaining candidate rows and matrices without
recomputation. Equations~\eqref{eq:candidate_weight_jackknife_fit}--
\eqref{eq:candidate_weight_shrinkage} then produce the stabilized weight $a_c$.

The pollutant-specific query prediction is
\begin{equation}
  \widehat v_{*,c}
  =\sum_{k=1}^{7}a_{c,k}\widehat v_{*,k,c}.
  \label{eq:supp_air_channel_combination}
\end{equation}

For $D=5$, the five held-out prompts and the complete prompt give $Q=D+1=6$
logical ICON evaluations per query.

\FloatBarrier
\section{Supplementary Results}\label{sec:extended}

The condition-level summaries in the main text and below make the scope and
limits of the comparisons explicit. They report conservation-law errors after one and 144
steps, every MFC task and component comparison, and every evaluated
air-quality lead time.

\subsection{Full out-of-distribution evaluation on canonical PDEs}

For conservation laws, the frozen ICON is pretrained on cubic fluxes and
CHOP(Conservation) is evolved on 100 trajectories from the $\sin u-\cos u$
flux. The single-step and 144-step evaluations each use 500 trajectories from
every displayed non-cubic flux. Table~\ref{tab:consv_singlestep} gives the exact
single-step arithmetic means marked in
main-text Fig.~\ref{fig:conservation_results}c and
adds the arithmetic mean error after the final rollout step.

\begin{figure}[!htbp]
  \centering
  \includegraphics[width=\linewidth]{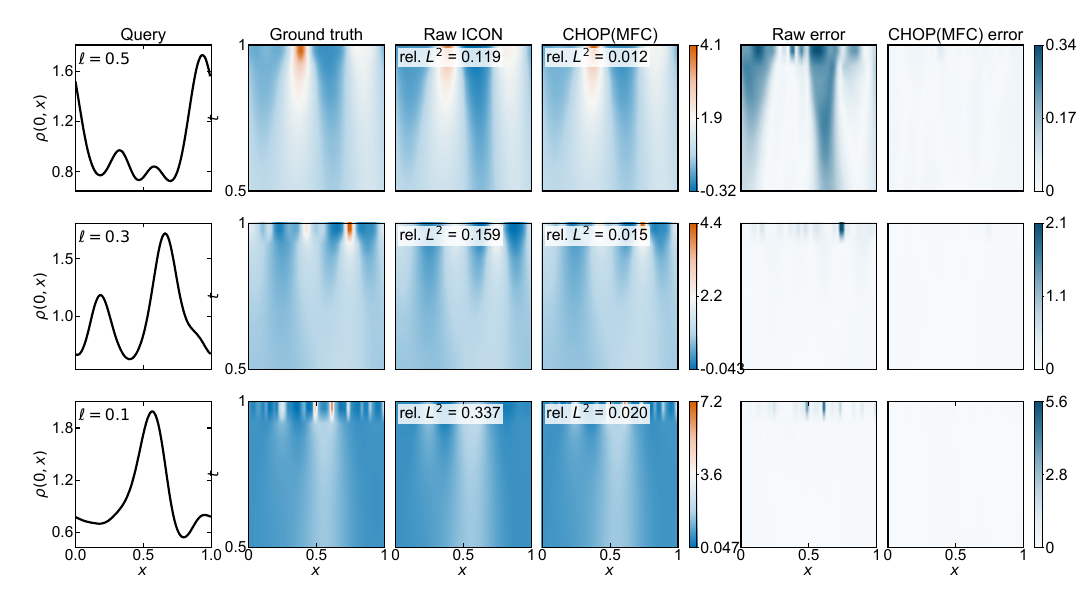}
  \caption{\textbf{Predictions for $\mathcal{M}_{g,\ell}^{1\to2}$.}
    Rows show $\ell=0.5$, $0.3$ and $0.1$; columns show the query, ground truth,
    Raw ICON and CHOP(MFC) predictions, followed by their pointwise absolute
    errors. Inline labels give relative $L^2$ errors. Within each row, the density
    heatmaps share one colour scale and the absolute-error heatmaps another.}
  \label{fig:app_mfc_gparam12_spacetime}
\end{figure}

\begin{figure}[!htbp]
  \centering
  \includegraphics[width=\linewidth]{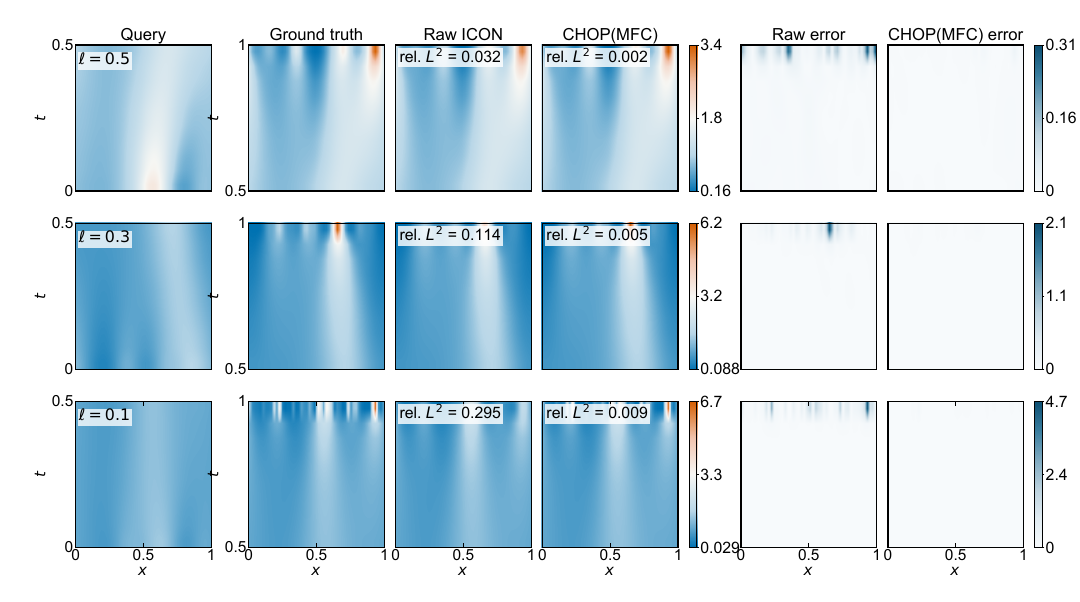}
  \caption{\textbf{Predictions for $\mathcal{M}_{g,\ell}^{2\to2}$.}
    Rows show $\ell=0.5$, $0.3$ and $0.1$; columns show the query, ground truth,
    Raw ICON and CHOP(MFC) predictions, followed by their pointwise absolute
    errors. Inline labels give relative $L^2$ errors. Within each row, the density
    heatmaps share one colour scale and the absolute-error heatmaps another.}
  \label{fig:app_mfc_gparam22_spacetime}
\end{figure}

For MFC, the frozen ICON is pretrained on the five operator families at
$\ell=1$, and CHOP(MFC) is evolved on 100 instances of
$\mathcal M_{g,0.5}^{2\to2}$. Evaluation uses 100 new instances from every
family at $\ell\in\{0.5,0.3,0.1\}$; fourteen of the resulting fifteen tasks are
absent from evolution. Table~\ref{tab:main} reports all fifteen numerical
  comparisons. Figures~\ref{fig:app_mfc_gparam12_spacetime} and
  \ref{fig:app_mfc_gparam22_spacetime} provide examples for the two families with
  spatiotemporal outputs that are absent from the qualitative panel in
  main-text Fig.~\ref{fig:mfc_results}d.

The final-step reduction exceeds the corresponding single-step reduction for
every flux. Buckley--Leverett moves from the smallest single-step reduction,
32.93\%, to the second-largest final-step reduction, 88.10\%.

All nine $\mathcal M_{g,\ell}$ conditions improve, and the reduction increases
within each family as $\ell$ decreases. Every key and value in these three
families is a density field. The $\mathcal M_{\rho,\ell}$ families instead use a
terminal-cost function as the key and a density field as the value, while
CHOP(MFC) continues to compute one affine center and scale from all visible key,
value and query samples. The complete program lowers no condition-level mean in
these six tasks. Because the rescaled base route, residual correction and outer
comparison act together, Table~\ref{tab:main} motivates the component comparison
below without assigning this outcome to one operation.

\subsection{Transfer across operator families and equations}

The cost-to-density experiments compare the complete program with a reduced
program that applies the retained $G_{\mathrm{corr}}$ to Raw ICON base
predictions. The frozen MFC ICON is
pretrained on all five operator families at $\ell=1$, while
$G_{\mathrm{corr}}$ is selected as part of CHOP(MFC) during evolution on 100
instances of $\mathcal M_{g,0.5}^{2\to2}$. We then evaluate CHOP(MFC) and
CHOP(Corr) on the same 100 held-out prompts for each cost-to-density family and
length scale.

CHOP(Corr) evaluates the complete prompt and each augmented held-out prompt
without affine rescaling, so the corresponding predictions require no inverse
rescaling. For each held-out example, it uses the same MFC prompt construction
as CHOP(MFC): the pair is removed, the remaining order is preserved and the
first remaining pair is appended so that ICON receives $D=5$ pairs. Raw ICON
supplies the resulting held-out base predictions and the complete-prompt base
prediction. CHOP(Corr) still computes the full-prompt scale
$\sigma_{\mathrm{sc}}$ from the visible key, value and query samples and
supplies $s_u=\sigma_{\mathrm{sc}}$ to Algorithm~\ref{alg:chop_G}. It retains
the internal gate in that algorithm and omits CHOP(MFC)'s outer route comparison.

Table~\ref{tab:mfc_component_ablation} places both programs beside their common
Raw ICON prediction. The complete program reproduces Raw ICON for all three
terminal-state conditions. For the spatiotemporal outputs, its relative $L^2$
error differs from Raw ICON on only three of the 300 queries: one decreases and
two increase, with the two increases raising the corresponding mean errors.
Applied directly to the Raw ICON route, $G_{\mathrm{corr}}$ lowers mean relative
$L^2$ error in all six conditions. The reduction is 1.80--4.60\% for
terminal-state prediction and 40.43--44.86\% for spatiotemporal prediction. The query-level
counts show the same difference in scale: CHOP(Corr) improves 45--61 of 100
terminal-state queries and 93--98 of 100 spatiotemporal queries.

\begin{table}[htbp]
  \centering
  \caption{\textbf{Cost-to-density component comparison.}
    Upper entries are mean $\pm$ sample s.d.\ relative $L^2$ error over the same
    $n=100$ prompts in each row.
    Lower-block reductions and count triplets compare each program with Raw ICON;
    negative reductions indicate increases.}
  \label{tab:mfc_component_ablation}
  \small
  \setlength{\tabcolsep}{4.0pt}
  \begin{tabular}{lcccc}
    \toprule
    Family & $\ell$ & Raw ICON & CHOP(MFC) & CHOP(Corr) \\
    \midrule
    \multirow{3}{*}{$\mathcal M_{\rho,\ell}^{1\to1}$}
      & 0.5 & $0.004637\pm0.001892$ & $0.004637\pm0.001892$ & $0.004554\pm0.001943$ \\
      & 0.3 & $0.004616\pm0.001821$ & $0.004616\pm0.001821$ & $0.004503\pm0.001850$ \\
      & 0.1 & $0.003976\pm0.001696$ & $0.003976\pm0.001696$ & $0.003793\pm0.001705$ \\
    \midrule
    \multirow{3}{*}{$\mathcal M_{\rho,\ell}^{1\to2}$}
      & 0.5 & $0.079546\pm0.031130$ & $0.080892\pm0.035557$ & $0.043864\pm0.018900$ \\
      & 0.3 & $0.074664\pm0.034394$ & $0.075392\pm0.036641$ & $0.043324\pm0.020183$ \\
      & 0.1 & $0.069952\pm0.025328$ & $0.069952\pm0.025328$ & $0.041670\pm0.017305$ \\
    \bottomrule
  \end{tabular}
  \par\medskip
  \begin{tabular}{lccccc}
    \toprule
    && \multicolumn{2}{c}{Mean reduction} & \multicolumn{2}{c}{Lower/equal/higher counts} \\
    \cmidrule(lr){3-4}\cmidrule(lr){5-6}
    Family & $\ell$ & CHOP(MFC) & CHOP(Corr) & CHOP(MFC) & CHOP(Corr) \\
    \midrule
    \multirow{3}{*}{$\mathcal M_{\rho,\ell}^{1\to1}$}
      & 0.5 & 0.00\% & 1.80\% & 0/100/0 & 45/34/21 \\
      & 0.3 & 0.00\% & 2.47\% & 0/100/0 & 48/31/21 \\
      & 0.1 & 0.00\% & 4.60\% & 0/100/0 & 61/25/14 \\
    \midrule
    \multirow{3}{*}{$\mathcal M_{\rho,\ell}^{1\to2}$}
      & 0.5 & $-1.69$\% & 44.86\% & 0/99/1 & 98/0/2 \\
      & 0.3 & $-0.97$\% & 41.98\% & 1/98/1 & 93/0/7 \\
      & 0.1 & 0.00\% & 40.43\% & 0/100/0 & 93/0/7 \\
    \bottomrule
  \end{tabular}
\end{table}

For cross-equation transfer, the complete CHOP(MFC) program evolved on
$\mathcal M_{g,0.5}^{2\to2}$ is coupled without further evolution to the
conservation-law ICON pretrained on cubic fluxes. All three methods in
Table~\ref{tab:crosstask} are evaluated on the same 500 trajectories for each
non-cubic flux. The transferred program lowers mean single-step relative $L^2$
error by 17.06--19.78\% on every flux after both the equation and frozen ICON
change. CHOP(Conservation), which was evolved on the
$\sin u-\cos u$ task and contains cyclic alignment and mass conservation, lowers
the same means by 32.93--50.68\%. The gap between the transferred and
conservation-law programs ranges from 13.15 to 33.62 percentage points.

\subsection{Air-quality transfer across eleven forecast lead times}


Tables~\ref{tab:air_forward_all_horizons} and
\ref{tab:air_reverse_all_horizons} report all eleven evaluated lead times. The
first eight lie within the 1 to 24\,h range used in pretraining, while 36, 48 and
72\,h extend beyond it. Each entry is the arithmetic mean of per-case RMSE
values defined in the same way as those plotted in
Main-text
Fig.~\ref{fig:air}c,d; the sample
unit is one pollutant field at one target
time.

The largest percentage reduction in mean per-case RMSE occurs at 1\,h in every
pollutant, transfer direction and example-selection combination. CHOP(Air)
lowers the mean in 77 of the 88 conditions. The eleven exceptions occur for
reverse PM$_{2.5}$ with similarity retrieval and for O$_3$ with random examples;
Tables~\ref{tab:air_forward_all_horizons} and
\ref{tab:air_reverse_all_horizons} give their lead times and magnitudes.

\begin{table}[H]
  \centering
  \caption{\textbf{RMSE for YRD$\to$BTHSA.}
    At lead time $h$, entries are arithmetic means over $n=8{,}760-h$ valid 2023
    target times. Each per-case RMSE is computed across the 228 BTHSA stations
    and reported in $\mu$g m$^{-3}$. Reductions are relative to Raw ICON.}
  \label{tab:air_reverse_all_horizons}
  \small
  \setlength{\tabcolsep}{3.2pt}
  \begin{tabular}{lrrrrrr}
    \toprule
    & \multicolumn{3}{c}{Random examples} & \multicolumn{3}{c}{Similarity retrieval} \\
    \cmidrule(lr){2-4}\cmidrule(lr){5-7}
    Lead time (h) & Raw & CHOP & Reduction (\%)
      & Raw & CHOP & Reduction (\%) \\
    \midrule
    \multicolumn{7}{l}{\textbf{PM$_{2.5}$}} \\
      1  & 23.96 &  8.64 & 64.0 & 21.76 &  8.94 & 58.9 \\
      2  & 23.80 & 12.61 & 47.0 & 21.59 & 12.83 & 40.6 \\
      3  & 23.81 & 15.54 & 34.7 & 21.65 & 15.21 & 29.7 \\
      4  & 23.96 & 17.79 & 25.7 & 21.90 & 16.86 & 23.0 \\
      5  & 24.21 & 19.56 & 19.2 & 22.28 & 18.11 & 18.7 \\
      6  & 24.55 & 20.95 & 14.7 & 22.74 & 19.13 & 15.9 \\
      12 & 26.97 & 25.73 &  4.6 & 25.79 & 23.31 &  9.6 \\
      24 & 29.93 & 29.71 &  0.7 & 28.48 & 27.95 &  1.9 \\
      36 & 33.02 & 33.00 &  0.1 & 31.74 & 31.68 &  0.2 \\
      48 & 34.53 & 34.20 &  1.0 & 32.72 & 33.11 & $-1.2$ \\
      72 & 36.28 & 35.33 &  2.6 & 34.15 & 34.19 & $-0.1$ \\
    \midrule
    \multicolumn{7}{l}{\textbf{O$_3$}} \\
      1  & 27.83 & 12.62 & 54.7 & 22.96 & 11.35 & 50.6 \\
      2  & 27.69 & 20.18 & 27.1 & 23.33 & 16.38 & 29.8 \\
      3  & 27.88 & 24.79 & 11.1 & 23.95 & 18.95 & 20.9 \\
      4  & 28.37 & 27.48 &  3.1 & 24.69 & 20.62 & 16.5 \\
      5  & 29.10 & 29.21 & $-0.4$ & 25.47 & 21.84 & 14.3 \\
      6  & 29.97 & 30.57 & $-2.0$ & 26.21 & 22.81 & 13.0 \\
      12 & 33.30 & 34.22 & $-2.8$ & 29.39 & 26.20 & 10.9 \\
      24 & 33.43 & 30.55 &  8.6 & 29.90 & 27.88 &  6.7 \\
      36 & 36.42 & 37.47 & $-2.9$ & 32.85 & 30.81 &  6.2 \\
      48 & 35.98 & 34.23 &  4.9 & 32.68 & 31.13 &  4.7 \\
      72 & 37.43 & 35.49 &  5.2 & 34.39 & 32.58 &  5.3 \\
    \bottomrule
  \end{tabular}
\end{table}

\begin{figure}[H]
  \centering
  \includegraphics[width=\linewidth]{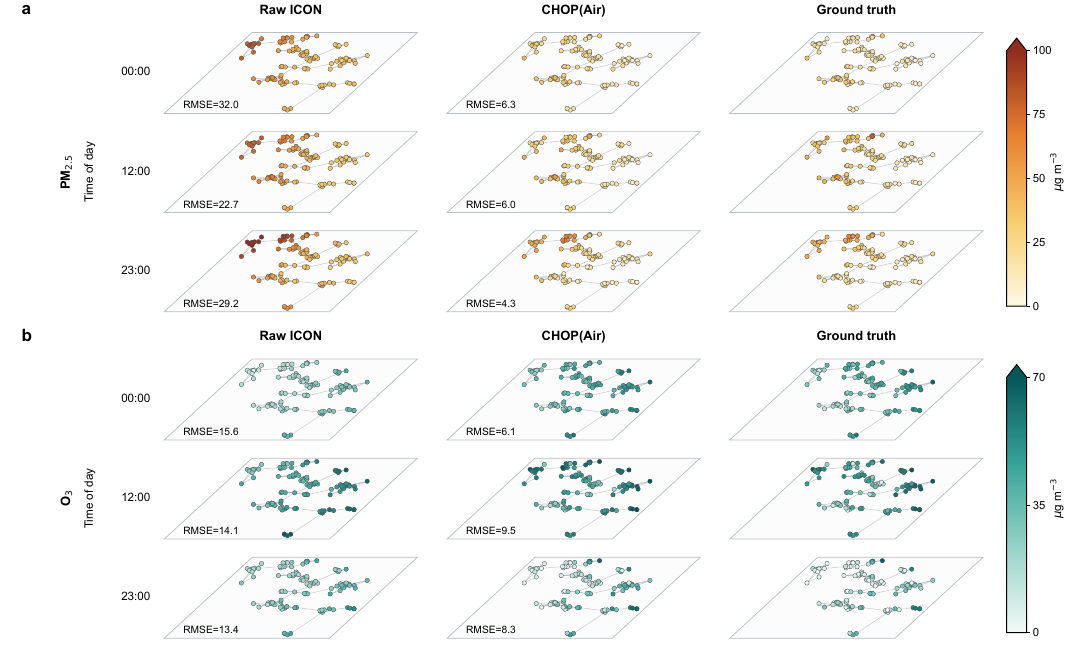}
  \caption{\textbf{Representative one-hour BTHSA$\to$YRD forecasts.}
    \textbf{a}, Raw ICON and CHOP(Air) PM$_{2.5}$ forecasts, with ground truth,
    at three target times on 3 February 2023.
    \textbf{b}, Corresponding O$_3$ forecasts. The query uses randomly selected
    YRD examples, and each RMSE is computed across the 127 YRD stations.
    Colour-bar arrowheads mark concentrations above the labelled maxima.}
  \label{fig:air_forecast_examples}
\end{figure}

\bibliographystyle{elsarticle-num}
\bibliography{main}

\end{document}